\documentclass[11pt]{article}

\usepackage[preprint]{acl}

\usepackage{times}
\usepackage{latexsym}

\usepackage{multirow}
\usepackage{algorithm}
\usepackage{algorithmic}
\usepackage[normalem]{ulem}
\useunder{\uline}{\ul}{}
\usepackage{pifont}
\usepackage[table,xcdraw]{xcolor}
\usepackage{amsmath}
\usepackage{subfigure}

\usepackage[T1]{fontenc}

\usepackage[utf8]{inputenc}

\usepackage{microtype}

\usepackage{inconsolata}

\usepackage{graphicx}

\title{RAAR: Retrieval Augmented Agentic Reasoning for Cross-Domain Misinformation Detection}

\author{%
  Zhiwei Liu\textsuperscript{1}\quad
  Runteng Guo\textsuperscript{2,3}\quad
  Baojie Qu\textsuperscript{2,3}\quad
  Yuechen Jiang\textsuperscript{1}\quad \\
  \textbf{Min Peng}\textsuperscript{2,3}\quad
  \textbf{Qianqian Xie}\textsuperscript{2,3}{\thanks{Corresponding Author}}\quad 
  \textbf{Sophia Ananiadou}\textsuperscript{1,2,3,4} \\ 
    \textsuperscript{1}The University of Manchester \quad 
    \textsuperscript{2}School of Artificial Intelligence, Wuhan University \quad  \\
    \textsuperscript{3}Center for Language and Information Research, Wuhan University \quad 
    \textsuperscript{4}ELLIS Manchester \quad  \\
\texttt{\{zhiwei.liu,sophia.ananiadou\}@manchester.ac.uk} \\
\texttt{xieq@whu.edu.cn}
}

\begin{document}
\maketitle
\begin{abstract}
Cross-domain misinformation detection is challenging, as misinformation arises across domains with substantial differences in knowledge and discourse. Existing methods often rely on single-perspective cues and struggle to generalize to challenging or underrepresented domains, while reasoning large language models (LLMs) though effective on complex tasks, are limited to same-distribution data. To address these gaps, we introduce RAAR, the first retrieval-augmented agentic reasoning framework for cross-domain misinformation detection. 
To enable cross-domain transfer beyond same-distribution assumptions, RAAR retrieves multi-perspective source-domain evidence aligned with each target sample’s semantics, sentiment, and writing style. To overcome single-perspective modeling and missing systematic reasoning, RAAR constructs verifiable multi-step reasoning paths through specialized multi-agent collaboration, where perspective-specific agents produce complementary analyses and a summary agent integrates them under verifier guidance. RAAR further applies supervised fine-tuning and reinforcement learning to train a single multi-task verifier to enhance verification and reasoning capabilities. 
Based on RAAR, we trained the RAAR-8b and RAAR-14b models. Evaluation on three cross-domain misinformation detection tasks shows that RAAR substantially enhances the capabilities of the base models and outperforms other cross-domain methods, advanced LLMs, and LLM-based adaptation approaches. The project will be released on https://github.com/lzw108/RAAR.
\end{abstract}

\section{Introduction}

Misinformation poses a growing threat across diverse domains such as public health, politics, finance, and science, where false claims frequently emerge in new contexts \cite{gradon2021countering}. In realistic scenarios, cross-domain misinformation detection involves training on known domains and predicting misinformation in unseen domains \cite{silva2021embracing}. Unlike general reasoning tasks, this setting requires not only effective feature learning and representation, but also an understanding of systematic cross-domain differences in knowledge structures and discourse characteristics \cite{doumas2022theory}. These differences induce substantial distribution shifts that render surface-level cues unreliable, making cross-domain misinformation detection a particularly challenging problem that demands reasoning over ambiguous and implicit evidence. Recent advances in reasoning capabilities have led to significant breakthroughs in a range of complex tasks \cite{sun2025survey}, suggesting strong potential for improving cross-domain generalization.

With the rise of large language models (LLMs), techniques such as few-shot learning \cite{singal2024evidence}, instruction tuning \cite{liu2024conspemollm,liu2025fmdllama}, and retrieval-augmented generation (RAG) \cite{niu2024veract} have been applied to misinformation detection, including adaptations of RAG and multi-agent approaches for cross-domain settings \cite{liu2025raemollm,li2025multi}. However, existing approaches have several limitations. First, ML and DL methods require separate fine-tuning or task-specific prompts for each misinformation task, meaning a single model can only handle one task and cannot generalize across tasks. Second, most methods analyze data from a single perspective, failing to integrate multiple dimensions such as sentiment, writing style, and semantics, which reduces effectiveness in complex or challenging domains. Third, systematic reasoning over ambiguous or implicit evidence is largely absent, further hindering performance in underrepresented domains. These limitations underscore the need for methods that can perform systematic, cross-domain reasoning over multidimensional and ambiguous evidence.
 
Moreover, reasoning LLMs such as GPT-o1 and DeepSeek-R1 \cite{guo2025deepseek} have been widely applied in domains such as finance \cite{qian2025fino1,liu2025fin}, mental health \cite{dai2025psyche, xiao2025mentrasuite}, and medicine \cite{chen2025towards}. However, these models primarily perform direct inference on data from the same distribution, focusing on information within a single domain. This makes it difficult to handle the substantial differences in knowledge structures and discourse encountered in cross-domain misinformation detection. Moreover, relying on a single model to construct reasoning paths limits their ability to exploit multidimensional, domain-specific information for complex tasks, constraining performance in unseen domains.


To address the research gaps, we propose RAAR, the first retrieval-augmented agentic reasoning framework for cross-domain misinformation detection, leveraging retrieval-augmented agentic reasoning with supervised fine-tuning (SFT) and reinforcement learning (RL), allowing cross-domain knowledge transfer, multidimensional reasoning, and multi-task optimization. RAAR incorporates three key stages to address the limitations of existing methods. (1) Retrieval augmented data building: to overcome the reliance on same-distribution assumptions and provide rich cross-domain evidence, we retrieve source-domain data similar in semantics, sentiment, and style for each target-domain sample, enabling multi-perspective reference for subsequent analysis. (2) Multi-Agent collaborated reasoning path building: to perform systematic, multidimensional reasoning over complex and ambiguous evidence, specialized sub-agents analyze semantic, sentiment, and stylistic aspects, while a summary agent integrates their outputs and conducts additional reasoning under verifier guidance. (3) Model optimization: to improve prediction accuracy and cross-domain reasoning, we apply SFT and RL, enabling a single model to handle multiple misinformation tasks (fake news, rumors, conspiracy theories) and overcoming the single-task limitation of traditional models.

Based on the RAAR framework, we trained RAAR-8b and RAAR-14b based on Qwen3-8b and Qwen3-14b, respectively. Both models achieve substantial improvements over the original base models and outperform other cross-domain methods, advanced LLMs, and LLM-adapted models. RAAR-14b surpasses the latest DeepSeek-V3.2-Reasoner in average F1 performance.

Our main contributions are as follows:

\begin{itemize}

\item We propose RAAR, the first complex reasoning framework for cross-domain misinformation detection, leveraging retrieval-augmented agentic reasoning with SFT and RL for cross-domain knowledge transfer and multi-step reasoning.

\item We employ multi-agent collaboration to integrate multiple perspectives into coherent reasoning paths, significantly enhancing cross-domain reasoning capabilities.

\item Experiments show that RAAR significantly boosts cross-domain misinformation detection, surpassing existing methods and advanced LLMs, and validating retrieval augmentation and multi-agent reasoning.

\end{itemize}

\section{Method}

\begin{figure*}[t]
\centering
  \includegraphics[width=2\columnwidth]{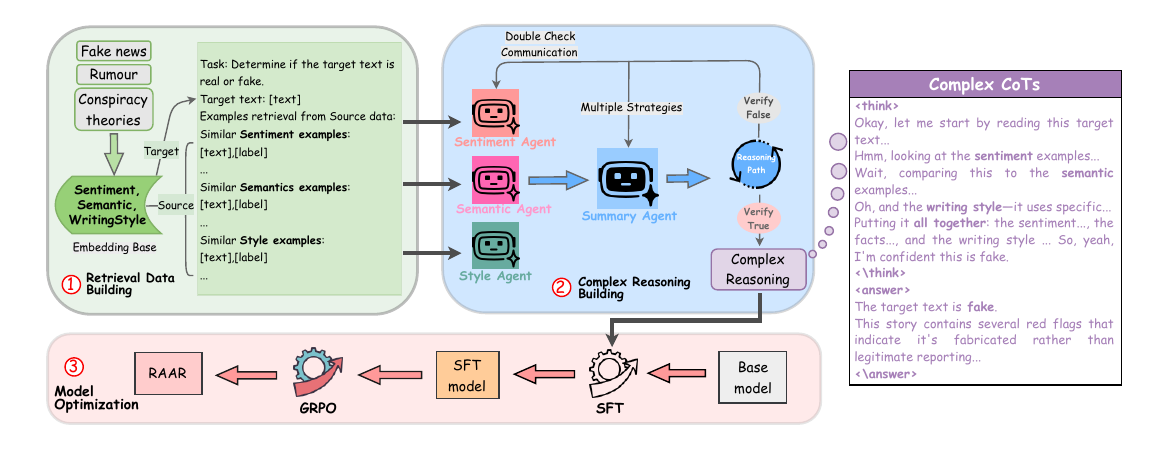}
  \caption{The architecture of RAAR. \ding{172} Retrieval Augmented Data Building: Retrieve source-domain samples similar to target-domain data in semantics, sentiment, and style for multi-perspective analysis. \ding{173} Multi-agent Collaborated Reasoning Path Building: Use multi-agent collaboration and a verifier to construct coherent, multi-perspective reasoning paths. \ding{174} Model Optimization: Fine-tune the model with SFT and RL to enhance cross-domain verification and reasoning.}
  \label{fig:mainmethod}
\end{figure*}


This section presents the RAAR framework, with its overall structure shown in Figure~\ref{fig:mainmethod}. Starting with task formulation, we then describe its three key stages in detail.

\subsection{Task Formulation}

Cross-domain misinformation detection aims to use labeled data from source domains to verify information in an unseen target domain. Specifically, given multiple source-domain datasets $D_s$ and a target-domain misinformation dataset $D_t$. The goal is to leverage source-domain knowledge to improve detection performance in a more challenging target domain. Formally, we define $\{D_s, D_t\} \subseteq D = \{ (x, y^*)\}$, where $x$ denotes a text content and $y^{*}$ is the corresponding label. Here, $D_t$ denotes the more challenging domains selected. The whole process includes three main stages. 

\textbf{Stage 1.} Retrieval Augmented Data Building: retrieve source-domain data similar in
semantics $\{(x_{sem}^s,y_{sem}^s)\}$, sentiment $\{(x_{sen}^s,y_{sen}^s)\}$, and style $\{(x_{sty}^s,y_{sty}^s)\}$ for each target-domain sample $\{(x^t,y^*)\}$:

\begin{equation}
\small
D_{ra} = Retrueval(D) = \{ (x_{ra}, y^{*})\},
\end{equation}

where $x_{ra}=\{x_t, \{(x_{sen}^s,y_{sen}^s)\},\{(x_{sem}^s,\\y_{sem}^s)\},\{(x_{sty}^s,y_{sty}^s)\}\}$. The $x_{ra}$ template can be found at Appendix \ref{app:template4LLMs}. $D_{ra}$ is split into two equal parts, $D_{search}$ and $D_{RL}$, which are used for reasoning path search in stage 2 and RL in stage 3, respectively.

\textbf{Stage 2.} Multi-agent Collaborated Reasoning Path Building: 
\begin{equation}
\small
\begin{split}
D_{SFT} = MultiAgents(D_{search}) = \{ (x_{ra}, \hat{CoT}, \hat{y}\},
\end{split}
\end{equation}
where $\hat{CoT}$ is the multi-perspective reasoning path, $\hat{y}$ is the final answer based on $\hat{CoT}$. 

\textbf{Stage 3:} Two-step model optimization: 
\begin{equation}
\small
\pi_{\theta}^{sft} = SFT_{(\pi_{\theta})}(\hat{CoT} , \hat{y} | x_{ra})
\end{equation}

\begin{equation}
\small
\hat{CoT}_{RL}, \hat{y}_{RL} = \pi_{\theta}^{sft}(D_{RL})
\end{equation}

\begin{equation}
\small
\pi_{\theta}^{RAAR} = RL_{(\pi_{\theta}^{sft})}(D_{RL}, \hat{CoT}_{RL}, \hat{y}_{RL}, Reward)
\end{equation}

where $\pi_{\theta}$ is an initial LLM models. The $reward$ includes accuracy and format. $\pi_{\theta}^{sft}$ acquires reasoning ability through SFT, then is fed $D_{RL}$ data to generate initial reasoning, which is further improved via RL.

After all stages, we obtain an optimized RAAR LLM $\pi_{\theta}^{RAAR}$, with the ultimate goal of improving $\pi_{\theta}$ to provide multi-perspective reasoning and a more accurate final answer $\hat{y}$. The whole process of the framework can be found at Algorithm \ref{alg:algorithm}.


\subsection{Stage 1: Retrieval Augmented Data Building \label{sec:stage1}}

This section focuses on retrieving multi-perspective reference samples from source-domain data for target-domain instances, serving as the foundation for subsequent reasoning construction. Previous studies have shown that sentiment, semantics, and writing style are important features for misinformation detection \cite{liu2025raemollm,bharadwaj2019fake,yang2024explore}. We further conduct sentiment analysis, semantic analysis, and writing style analysis in the Appendix \ref{app:sssanalysis}. Results show that examples retrieved from the source domain based on these features are likely to share the same labels as those in the target domain, making them more useful for the subsequent reasoning process to analyze commonalities across domains. Meanwhile, we conducted statistical analyses to compare the performance of different models and ultimately selected Emollama \cite{liu2024emollms}, Roberta \cite{liu2019roberta}, and StyleEmb \cite{wegmann2022same} as our embedder. We encode the input using EmoLLama \cite{liu2024emollms}, RoBERTa\cite{liu2019roberta}, and StyleEmb \cite{wegmann2022same}, and extract the last hidden state as embeddings for sentiment ($E_{sen}^t$, $E_{sen}^s$), semantic ($E_{sem}^t$, $E_{sem}^s$), and writing style ($E_{sty}^t$, $E_{sty}^s$), respectively. We next introduce the sentiment-based retrieval process.


\begin{equation}
\small
E_{sen}^t = EmoLlama(D_t) = (e_{sen}^{t_1}, e_{sen}^{t_2}, ..., e_{sen}^{t_M})
\end{equation}

\begin{equation}
\small
E_{sen}^s = EmoLlama(D_s) = (e_{sen}^{s_1}, e_{sen}^{s_2}, ..., e_{sen}^{s_N})
\end{equation}





where $M$ is the number of target domain items, $N$ is the number of source domain items. Next, we traverse the sentiment embeddings of the target domain ($e_{sen}^t$) within $E_{sen}^t$ and compute their cosine similarity with each sentiment embedding $e_{sen}^s$ from the source domain $E_{sen}^s$. For each target-domain instance, we then select the top-$k$ most similar source-domain examples ($\{(x_{sen}^s,y_{sen}^s)\}$) based on these similarity scores to construct $D_{ra}$, which serves as the few-shot examples for the subsections. The retrieval procedure for semantic and style embeddings follows the same methodology as for sentiment. Finally, we obtain the complete $D_{ra}=\{(x_{ra},y^*)\}$. $x_{ra}=\{x_t, \{(x_{sen}^s,y_{sen}^s)\},\{(x_{sem}^s,y_{sem}^s)\},\{(x_{sty}^s,y_{sty}^s)\}\}$ denotes the retrieval augmented items and $y^*$ denotes the ground-truth answer. 

We divide $D_{ra}$ equally into $D_{search}$ and $D_{RL}$, where $D_{search}$ is used for reasoning path search as detailed in Section \ref{sec:complexreasoning}, and $D_{RL}$ is used for the reinforcement learning component described in Section \ref{sec:RL}.

\subsection{Stage 2: Multi-agent Collaborated Reasoning Path Building \label{sec:complexreasoning}}



After obtaining $D_{search}$, we introduce a multi-agent framework to enable systematic, multidimensional reasoning over complex and ambiguous evidence. Three specialized SubAgents, namely the Sentiment Agent ($A_{sen}$), Semantic Agent ($A_{sem}$), and Style Agent ($A_{sty}$), analyze $D_{search}$ from their respective perspectives and generate dimension-specific responses with supporting reasons. A Summary Agent ($A_{sum}$) then integrates these outputs to construct coherent reasoning paths.


\begin{equation}
\small
\begin{split}
CoT^{sub}, y^{sub} = SubAgents(D_{search}) = \\\{{A_{sen}(x_t,\{(x_{sen}^s,y_{sen}^s)\})}, {A_{sem}(x_t,\{(x_{sem}^s,y_{sem}^s)\})}, \\{A_{sty}(x_t,\{(x_{sty}^s,y_{sty}^s)\})}\}
\end{split}
\end{equation}

where $CoT^{sub} = \{CoT_{sen}, CoT_{sem}, CoT_{sty}\}$ and $y^{sub} = \{y_{sen}, y_{sem}, y_{sty}\}$. Subsequently, the Summary Agent ($A_{sum}$) integrates these perspectives and performs additional reasoning to produce the initial reasoning chain ($CoT_0$) and prediction ($y_0$).

\begin{equation}
\small
CoT_{0}, y_{0} = A_{sum}(CoT^{sub}, y^{sub})
\end{equation}

\begin{equation}
\small
Verifier(y_{0},y^*) \in \{True, False\}
\end{equation}

An LLM-based verifier is employed to determine whether $y_0$ is correct. If not, the three sub-agents and the summarizing agent will each adopt multiple strategies collaboratively to identify the correct reasoning path.

Strategy for SubAgents:
\begin{itemize}
    \item \textbf{Double-Check (DC)}: This strategy informs the agent that the summarizing agent’s answer is incorrect. The agent needs to conduct self-verification. If an error is found, it should correct it. If it maintains its original stance, it should provide additional justification.

    \item \textbf{Communication}: Under this strategy, each agent is informed of the other agents’ answers and reasoning. Similarly, if any issues are identified, the agent should make corrections; if it maintains its position, it should offer further justification.
\end{itemize}

Based on existing studies~\cite{chen2025towards,qian2025fino1}, we improved and adapted the following strategies for Summary Agent: 
\textbf{(1) Cross-agent consolidation}: Aggregate sub-agent outputs to produce a unified answer.
\textbf{(2) Cross-agent reconsideration}: Re-evaluate the current reasoning by cross-checking sub-agent rationales.
\textbf{(3) Cross-agent diversification}: Leverage sub-agent signals to explore an alternative reasoning path.
\textbf{(4) Cross-agent verification}: Assess the consistency of the current reasoning across sub-agent perspectives.
\textbf{(5)Cross-agent rectification}: Identify and correct potential flaws in the current reasoning using sub-agent contents.

The above strategies ensure not only efficiency in identifying the correct reasoning path but also sufficient diversity in the reasoning process. Due to cost and time considerations, we set the total number of search rounds to $N=3$. During the exploration process, updates are first made from the summary agent. If the answer is still incorrect, update then backtrack to the subagents\footnote{It is worth noting that to avoid getting stuck in a loop, when all three agents provide the same incorrect answer, we first update the subagents to generate as diverse answers as possible for the summary agent to reference.}. If the correct reasoning path is not found after more than N iterations, we provide appropriate hints to force each agent to revisit previous steps, refine its hypotheses based on the verifier's feedback, and adjust its reasoning process to arrive at the correct reasoning path. After obtaining the complete and correct reasoning paths, we use LLMs to organize all the reasoning steps and generate a more human-like, intuitive thinking process ($\hat{CoT}$). 

\begin{equation}
\small
\hat{CoT} = LLM{rephrase}(CoT_0, CoT_1,...CoT_N, x_{ra})
\end{equation}

Based on this natural reasoning, generate a rich and high-quality final response ($\hat{y}$), where $\hat{y}$ consists of a clear answer in the first paragraph, followed by a summarized explanation in the second paragraph.

\begin{equation}
\small
\hat{y} = LLM{refine}(\hat{CoT}, x_{ra})
\end{equation}

All LLMs involved in this section use DeepSeek-v3.2-chat due to its cost advantage. The prompts for all agents, as well as those for all steps involving LLMs, can be found in the Appendix \ref{app:prompts4reasonpaths}.

\subsection{Stage 3: Model Optimization \label{sec:stage3}}

\subsubsection{Retrieval Augmented SFT}

After collecting the human-like $\hat{CoT}$s and the final answers $\hat{y}$, we use the template in Appendix \ref{app:template4LLMs} to construct the dataset ($D_{SFT}$) for SFT. The input is the retrieval augmented data $x_{ra}$. Given an initial LLM model $\pi_{\theta}$. We can get the $\pi_{\theta}^{sft}$ through SFT.

\begin{equation}
\small
\begin{split}
L_{SFT}(\theta) \gets -log \pi_{\theta}(\hat{CoT}, \hat{y} \mid x_{ra}) \\
\theta \gets \theta -\eta \nabla(\theta)L_{SFT}(\theta)
\end{split}
\end{equation}


The SFT stage guides the model to systematically construct its reasoning process before producing the final answer. Training is performed on retrieval-augmented data that encourages multi-step, multi-perspective reasoning, with target outputs consisting of a final answer accompanied by a summarizing explanation. This approach fosters a coherent analytical process, enhancing both interpretability and accuracy in cross-domain misinformation detection.

\subsubsection{RL \label{sec:RL}}

In the RL stage, we adopt the Group Relative Policy Optimization (GRPO) algorithm \cite{Shao2024DeepSeekMathPT} to further enhance reasoning ability of $\pi_{\theta}^{sft}$. Each sample in the training dataset $D_{RL}$ consists of two components $(x_{ra},y^*)$, where $x_{ra}$ represents the question and $y^*$ denotes the model’s output containing only the final answer without the reasoning process. Similar to previous studies~\cite{zhu2025dianjin,liu2025fin}, two reward mechanisms are employed: an accuracy reward, which encourages the model to produce correct answers, and a format reward, which ensures that the generated outputs adhere to the expected format.

\begin{itemize}
    \item \textbf{Format reward:} We introduce a format reward to ensure that the model’s output follows the structure: ``<think>\textbackslash n\textit{[reasoning]}\textbackslash n</think>\textbackslash n\textbackslash n<answer> \textbackslash n\textit{[firstsentence]}\textbackslash n\textbackslash n\textit{[response]}</answer>''. $\textit{[firstsentence]}$ contains the explicit predicted label. If the output conforms to this format, the model receives a reward of 1; otherwise, the reward is 0.
    \item \textbf{Accuracy reward:} The reward is 1 for a correct answer, 0.1 for incorrect, and 0 for None. Unlike LLM-based evaluation, we compute it directly using hard-coded rules, requiring the model to give the answer in the first sentence.

\end{itemize}

\section{Experiments}

\begin{table*}[t]
\footnotesize
\resizebox{1\textwidth}{!}{
\begin{tabular}{lccccccccccccccc}
\hline
           & \multicolumn{4}{c}{AMTCele}                                       &           & \multicolumn{4}{c}{PHEME}                                         &           & \multicolumn{4}{c}{COCO}                                          &                \\ \cline{2-5} \cline{7-10} \cline{12-15}
Models     & ACC            & Pre            & Recall         & F1             &           & ACC            & Pre            & Recall         & F1             &           & ACC            & Pre            & Recall         & F1             & Ave. F1        \\ \hline
           & \multicolumn{15}{c}{\textit{\textbf{cross-domain methods}}}                                                                                                                                                                                                \\ \cline{2-16} 
MDFEND     & 0.642          & 0.663          & 0.642          & 0.630          &           & 0.433          & {\ul 0.622}    & 0.575          & 0.420          &           & 0.720          & 0.722          & 0.640          & 0.662          & 0.570          \\
EDDFN      & 0.680          & 0.686          & 0.680          & 0.678          &           & 0.331          & 0.536          & 0.507          & 0.278          &           & 0.715          & 0.681          & 0.680          & 0.680          & 0.545          \\
MOSE       & 0.492          & 0.248          & 0.492          & 0.330          &           & 0.555          & 0.546          & 0.554          & 0.534          &           & 0.466          & 0.355          & 0.343          & 0.348          & 0.404          \\
CANMD      & 0.504          & 0.751          & 0.504          & 0.342          &           & 0.482          & 0.570          & 0.569          & 0.481          &           & 0.628          & 0.595          & 0.583          & 0.579          & 0.468          \\
MetaAdapt  & 0.608          & 0.576          & 0.820          & 0.677          &           & \textbf{0.692} & \textbf{0.625} & 0.508          & 0.433          &           & 0.511          & 0.492          & 0.504          & 0.487          & 0.532          \\ \hline
           & \multicolumn{15}{c}{\textit{\textbf{ZS}}}                                                                                                                                                                                                          \\ \cline{2-16} 
gpt-4.1    & 0.856          & 0.862          & 0.856          & 0.855          &           & 0.579          & 0.524          & 0.525          & 0.524          &           & 0.781          & 0.596          & 0.537          & 0.537          & 0.639          \\
gpt-5-mini & 0.780          & 0.780          & 0.780          & 0.780          &           & {\ul 0.586}    & 0.549          & 0.554          & 0.547          &           & 0.765          & 0.798          & 0.686          & 0.691          & 0.673          \\
Deepseek-R & 0.832          & 0.833          & 0.832          & 0.832          &           & 0.512          & 0.515          & 0.518          & 0.495          &           & 0.776          & 0.810          & 0.705          & 0.708          & 0.679          \\
Deepseek-C & 0.846          & 0.856          & 0.846          & 0.845          &           & 0.496          & 0.503          & 0.504          & 0.481          &           & 0.712          & 0.747          & 0.609          & 0.612          & 0.646          \\ 
Llama-8b   & 0.666          & 0.554          & 0.444          & 0.489          &  & 0.462          & 0.344          & 0.315          & 0.309          &  & 0.471          & 0.463          & 0.340          & 0.358          & 0.385          \\

Qwen-8b    & 0.776          & 0.779          & 0.776          & 0.775          &           & 0.571          & 0.361          & 0.361          & 0.358          &           & 0.641          & 0.666          & 0.529          & 0.516          & 0.550          \\
Qwen-14b   & 0.742          & 0.765          & 0.742          & 0.736          &           & 0.567          & 0.375          & 0.380          & 0.365          &           & 0.706          & 0.726          & 0.604          & 0.610          & 0.570          \\
Qwen-32b   & 0.742          & 0.757          & 0.742          & 0.738          &           & 0.560          & 0.548          & 0.556          & 0.537          &           & 0.687          & 0.729          & 0.575          & 0.566          & 0.614          \\ \hline
           & \multicolumn{15}{c}{\textit{\textbf{RA}}}                                                                                                                                                                                                          \\ \cline{2-16} 
gpt-4.1    & 0.820          & 0.847          & 0.820          & 0.816          &           & 0.570          & 0.517          & 0.518          & 0.516          &           & 0.812          & 0.817          & 0.759          & 0.755          & 0.696          \\
gpt-5-mini & 0.790          & 0.792          & 0.790          & 0.790          &           & 0.584          & 0.553          & 0.560          & {\ul 0.550}    & {\ul }    & 0.786          & 0.807          & 0.714          & 0.725          & 0.688          \\
Deepseek-R & \textbf{0.868} & \textbf{0.871} & \textbf{0.868} & \textbf{0.868} & \textbf{} & 0.531          & 0.522          & 0.526          & 0.509          &           & 0.793          & {\ul 0.818}    & 0.726          & 0.739          & {\ul 0.705}    \\
Deepseek-C & {\ul 0.862}    & {\ul 0.864}    & {\ul 0.862}    & {\ul 0.862}    & {\ul }    & 0.518          & 0.518          & 0.521          & 0.500          &           & 0.763          & 0.797          & 0.684          & 0.689          & 0.684          \\
Llama-8b   & 0.820          & 0.549          & 0.547          & 0.548          &  & 0.448          & 0.339          & 0.339          & 0.298          &  & 0.598          & 0.592          & 0.586          & 0.568          & 0.472          \\
Qwen-8b    & 0.786          & 0.799          & 0.786          & 0.784          &           & 0.526          & 0.356          & 0.359          & 0.341          &           & 0.716          & 0.712          & 0.626          & 0.634          & 0.586          \\
Qwen-14b   & 0.782          & 0.785          & 0.782          & 0.781          &           & 0.535          & 0.540          & 0.547          & 0.520          &           & 0.750          & 0.753          & 0.666          & 0.677          & 0.659          \\
Qwen-32b   & 0.804          & 0.804          & 0.804          & 0.804          &           & 0.535          & 0.531          & 0.536          & 0.515          &           & 0.735          & 0.750          & 0.652          & 0.666          & 0.662          \\ \hline
           & \multicolumn{15}{c}{\textit{\textbf{Adaptation}}}                                                                                                                                                                                                  \\ \cline{2-16} 
IT\_zs     & 0.784          & 0.794          & 0.784          & 0.782          &           & 0.461          & 0.568          & 0.561          & 0.459          &           & \textbf{0.819} & 0.801          & \textbf{0.785} & \textbf{0.792} & 0.678          \\
IT\_ra     & 0.750          & 0.509          & 0.500          & 0.503          &           & 0.578          & 0.578          & \textbf{0.591} & \textbf{0.562} & \textbf{} & 0.660          & 0.552          & 0.572          & 0.530          & 0.532          \\
RAEmoLLM   & 0.808          & 0.843          & 0.808          & 0.803          &           & 0.570          & 0.527          & 0.530          & 0.525          &           & {\ul 0.812}    & \textbf{0.822} & {\ul 0.761}    & 0.758          & 0.696          \\
MARO       &  0.688              &        -        &        -        &  0.680              &           &       -         &       -         &   -             &        -        &           &      -          &       -         &         -       &         -       &        -        \\
RAAR-8b    & 0.812          & 0.813          & 0.812          & 0.812          &           & 0.536          & 0.558          & 0.567          & 0.528          &           & 0.776          & 0.779          & 0.707          & 0.727          & 0.689          \\
RAAR-14b   & 0.842          & 0.842          & 0.842          & 0.842          &           & 0.541          & 0.572          & {\ul 0.582}    & 0.535          &           & 0.799          & 0.792          & 0.748          & {\ul 0.759}    & \cellcolor[HTML]{FCFF2F}\textbf{0.712} \\ \hline
\end{tabular}
}
\caption{Main results on three datasets. ``ZS'' denotes the zero-shot setting. ``RA'' denotes few-shot setting based retrieval augmented data. The results of MARO in AMTCele are refer to \cite{li2025multi} \label{tab:mainresults}}
\end{table*}


\subsection{Datasets}

We conduct experiments on AMTCele \cite{perez2018automatic1}, PHEME \cite{kochkina2018all2}, and COCO \cite{langguth2023coco3}. AMTCele, a cross-domain fake news dataset, includes FakeNewsAMT and Celebrity with both legit and fake labels. PHEME is a Twitter dataset containing rumours and non-rumours across nine breaking news events. COCO is a conspiracy-oriented Twitter dataset covering 12 conspiracy categories, with tweets labeled as Unrelated, Conspiracy, or Related. We assess the difficulty of each domain in AMTCele and PHEME using GPT-4.1 and DeepSeek-Chat to select relatively challenging domains as the test set. Detailed evaluation, analysis, and statistics are provided in Appendix~\ref{app:dataprocessstatistics}.

\subsection{Baselines}

\textbf{1) LLMs:} We select mainstream LLMs as our baseline models for comparison, including reasoning-oriented models such as GPT-5-mini \cite{openai_gpt5_mini}, DeepSeek-v3.2-Reasoner \cite{deepseek_models}, and Qwen3 (8B, 14B, and 32B) \cite{qwen3_techreport}. In addition, we include advanced LLMs, including LLama3.1-8b-instruct (Llama-8b) \cite{dubey2024llama}, GPT-4.1 \cite{openai_gpt41}, DeepSeek-v3.2-Chat \cite{deepseek_models}, as strong baselines. In this paper, we consider two experimental settings: zero-shot evaluation and few-shot evaluation enhanced by RAAR retrieval of in-context examples.

\textbf{2) Cross-domain methods:} MOSE \cite{qin2020multitask}: mixture-of-experts (MoE). EDDFN \cite{silva2021embracing}: multimodal features. MDFEND \cite{nan2021mdfend} Domain Gate mechanism based on MoE framework. CANMD \cite{yue2022contrastive}: contrastive learning. MetaAdapt \cite{yue2023metaadapt}: meta-learning paradigm.

\textbf{3) LLM-based Adaptation methods:} \textbf{Instruction Tuning}: We fine-tune two variants of the model based on Qwen3-8b, including a zero-shot fine-tuned version (IT\_zs) (like conspemollm \cite{liu2024conspemollm}) and a retrieval-augmented fine-tuned version with few-shot examples (IT\_ra). \textbf{RAEmoLLM} \cite{liu2025raemollm}: RAG. \textbf{MARO} \cite{li2025multi}: Multi agents.

Details of the above baselines can be found in the Appendix \ref{app:baselines}. The templates for zero-shot, few-shot based on RA data, and instruction-tuning methods can be found at Appendix \ref{app:template4LLMs}.

\subsection{Settings and Evaluation}

We develop two LLMs based on the RAAR framework, namely RAAR-8B and RAAR-14B, which are built upon the reasoning versions of Qwen3-8B and Qwen3-14B, respectively. In Stage 1, we select the top two examples for each dimension, details are provided in Section \ref{sec:ablation}. Details of SFT and RL settings can be found in Appendix \ref{app:settings}.

Following previous methods, we evaluate model performance using multiple metrics, including accuracy, precision, recall, and F1 score (All metrics use the macro variant).

\subsection{Main results}

Table \ref{tab:mainresults} presents the performance of all baselines and our methods. 
Overall, RAAR-14b achieves the highest average F1 performance across the three datasets. Moreover, RAAR substantially improves the performance of the base models (i.e., from Qwen3-8b to RAAR-8b, and from Qwen3-14b to RAAR-14b), and outperforms other cross-domain methods, advanced LLMs, and LLMs-based adaptation methods, demonstrating the effectiveness of the RAAR framework for complex cross-domain misinformation detection tasks. Next, we will conduct a detailed analysis.

\textbf{RAAR outperforms existing LLM-based adaptation methods:} The final part of the table compares LLM-based adaptation methods.
\textbf{(1) Dataset-specific adaptation:} Instruction-tuning approaches tend to overfit the datasets they are trained on. For example, IT\_zs excels on COCO while IT\_fs performs best on PHEME, yet both methods show only average performance on the remaining datasets, particularly on the small-scale AMTCele. This highlights that fine-tuning adaptation without reasoning may not generalize well to datasets with different scales or domain characteristics. 
\textbf{(2) RAG-based adaptation:} RAEmoLLM demonstrates an overall improvement compared to GPT-4.1 in the zero-shot setting, yet it still lags behind RAAR-14b. This suggests that augmenting retrieval with sentiment alone is insufficient for complex cross-domain misinformation detection. Without mechanisms for systematic multi-step reasoning or integration of complementary perspectives, RAEmoLLM cannot fully capture ambiguous or implicit evidence, limiting its ability to generalize to diverse domains.
\textbf{(3) MultiAgent-based adaptation:} Despite leveraging multiple agents, MARO relies on simple analysis and lacks systematic multi-perspective reasoning, with rule-based constraints limiting its applicability to the COCO propagation intent detection task. As a result, it shows limited effectiveness on challenging or underrepresented domains, where ambiguity and implicit cues require integration of sentiment, stylistic, and semantic information. \textbf{Overall}, these observations further highlight the superiority of the RAAR framework, which integrates retrieval-augmented multi-agent collaborative reasoning with two stage: sft-rl model optimization to achieve robust cross-domain performance.

\textbf{Comparison with LLMs:} 
\textbf{(1) Larger models perform better:} Results show that large-scale models (GPT and DeepSeek families) consistently outperform smaller models (Qwen series and Llama-8B) in cross-domain misinformation detection, reflecting advantages in representation capacity and world knowledge that enable more robust reasoning under distribution shifts.
\textbf{(2) Few-shot learning yields clear gains:}
Few-shot results consistently outperform zero-shot performance, indicating that retrieval-augmented in-context examples provide effective cross-domain cues and enable more reliable reasoning.
\textbf{(3) RAAR outperforms advanced LLMs:} RAAR framework substantially strengthens base models, with RAAR-14B achieving higher average F1 scores than advanced proprietary models such as GPT-4.1 and the DeepSeek series, demonstrating its effectiveness in cross-domain settings.
\textbf{(4) Reasoning models excel on complex tasks.}
The reasoning version of DeepSeek outperforms the chat version in most cases, suggesting that explicit reasoning capabilities are crucial for handling complex cross-domain scenarios.
\textbf{(5) Special case:} Notably, under the zero-shot setting, the DeepSeek series and GPT-4.1 perform particularly well on the AMTCele dataset. Consistent with \cite{liu2025raemollm}, a possible explanation is that AMTCele is collected from public websites and may overlap with the training corpora of these models, enabling them to better exploit such information.

\textbf{ RAAR outperforms existing cross-domain methods:} We can observe that the RAAR series consistently outperforms cross-domain methods on complex domain tasks across all three benchmarks, highlighting the advantages of LLMs and the proposed training strategy. In contrast, existing cross-domain methods exhibit highly unstable performance, even when trained separately on each dataset. For example, MDFEND, EDDFN, and MetaAdapt perform well on AMTCele and COCO, but perform poorly on PHEME, possibly due to their limited ability to understand short texts. Meanwhile, MOSE and CANMD show unsatisfactory performance on AMTCele, indicating their inability to effectively handle unstructured long-text data.


\subsection{Ablation study \label{sec:ablation}}

\begin{table}[]
\resizebox{0.48\textwidth}{!}{
\begin{tabular}{lccccccccc}
\hline
\multicolumn{1}{c}{} & \multicolumn{2}{c}{AMTCele}     &  & \multicolumn{2}{c}{PHEME}       &  & \multicolumn{2}{c}{COCO}        & \multicolumn{1}{l}{}                                          \\ \cline{2-3} \cline{5-6} \cline{8-9}
Models               & ACC            & F1             &  & ACC            & F1             &  & ACC            & F1             & Ave. F1                                                       \\ \hline
RAAR-8b              & {\ul 0.812}    & {\ul 0.812}    &  & 0.536          & 0.528          &  & 0.776          & 0.727          & \cellcolor[HTML]{FFFE65}{\color[HTML]{000000} \textbf{0.689}} \\
Reason\_zs            & 0.786          & 0.785          &  & \textbf{0.554} & \textbf{0.545} &  & 0.763          & 0.535          & 0.622                                                         \\
w/o RA               & 0.748          & 0.743          &  & 0.525          & 0.520          &  & 0.756          & 0.710          & 0.658                                                         \\
1shot RA             & 0.778          & 0.777          &  & {\ul 0.544}    & {\ul 0.539}    &  & 0.773          & 0.725          & 0.680                                                         \\
4shot RA             & 0.804          & 0.803          &  & 0.512          & 0.510          &  & \textbf{0.790} & \textbf{0.746} & 0.686                                                         \\ \hline
w/o agents           & 0.778          & 0.777          &  & 0.518          & 0.514          &  & 0.783          & 0.739          & 0.677                                                         \\
w/o semantic         & 0.790          & 0.790          &  & 0.531          & 0.526          &  & {\ul 0.784}    & {\ul 0.740}    & 0.685                                                         \\
w/o sentiment        & 0.786          & 0.786          &  & 0.538          & 0.531          &  & 0.775          & 0.727          & 0.681                                                         \\
w/o style            & {\ul 0.810}    & 0.810          &  & 0.507          & 0.335          &  & 0.775          & 0.723          & 0.623                                                         \\ \hline
w/o RL               & {\ul 0.810}    & 0.810          &  & 0.525          & 0.515          &  & 0.764          & 0.712          & 0.679                                                         \\
SFT+Full             & 0.782          & 0.782          &  & 0.535          & 0.522          &  & 0.761          & 0.718          & 0.674                                                         \\
SFT+PPO              & 0.786          & 0.784          &  & 0.506          & 0.498          &  & 0.774          & 0.727          & 0.670                                                         \\
RAAR-Llama           & \textbf{0.834} & \textbf{0.834} &  & 0.506          & 0.504          &  & 0.773          & 0.726          & {\ul 0.688}                                                   \\ \hline
\end{tabular}
}
\caption{Ablation Analysis. “w/o” denotes without. Bold indicates best performance. Underlining denotes the second-best performance. \label{tab:ablationstudy}}
\end{table}

To evaluate the contribution of each RAAR module, we construct variant models and conduct ablation experiments by systematically removing individual components (all experiments are based on Qwen3-8B).
\textbf{(1) For Retrieval Data Building:}
\textbf{Reason\_zs:} Without retrieval-augmented examples, reasoning is performed solely based on the target text. 
\textbf{w/o RA:} Without using retrieval augmentation, few-shot examples are randomly sampled. 
\textbf{1shot RA:} Retrieval-augmented with one example for each dimension. 
\textbf{4shot RA:} Retrieval-augmented with four examples for each dimension.
\textbf{(2) For Complex Reasoning Building:}
\textbf{w/o agents:} Without using any agents, reasoning is performed directly based on the retrieval-augmented data.
\textbf{w/o semantic:} Without using the semantic agent.
\textbf{w/o sentiment:} Without using the sentiment agent.
\textbf{w/o style:} Without using the writing style agent.
\textbf{(3) For Training:}
\textbf{w/o RL:} Without the RL step, SFT is performed only using $D_{SFT}$.
\textbf{SFT+Full:} Conduct SFT using the complex path constructed from full training datasets.
\textbf{SFT+PPO:} The RL phase is trained using PPO.
\textbf{RAAR-Llama:} Apply RAAR to train Llama-8b.

Table \ref{tab:ablationstudy} presents the ablation results. For the average F1 score, all variants show varying degrees of performance decline. For the \textit{Reason\_zs} variant, similar to the previously mentioned instruction-tuning, it only excels on a single dataset and is prone to overfitting to a specific type of data. \textit{w/o RA} results show that randomly selected few-shot examples lead to a relatively large performance drop, demonstrating the importance of retrieval augmentation for obtaining effective examples. By comparing the results of \textbf{1shot} and \textbf{4shot} settings, we find that providing more examples generally improves performance. However, the number of examples should be balanced, as too many can also make it difficult for the model to understand.

For the Complex Reasoning Building section, all w/o agent variants show little impact on the COCO dataset, and in some cases even perform slightly better. A possible reason is that COCO consists entirely of COVID-19–related content; although the topics are subdivided, the sentiment, semantics, and writing style across different topics are likely quite similar, making it difficult for agents to differentiate during analysis. For datasets with more distinct topics, such as AMTCele and PHEME, the results indicate that each agent is highly valuable. Notably, the \textit{w/o} sentiment variant on AMTCele drops by 0.26, demonstrating the importance of sentiment analysis for long-text news\footnote{The \textit{w/o style} variant performs poorly on PHEME, likely because during testing the model generates repeated and irrelevant words across multiple samples, preventing it from producing correct answers.}.

For the training part, the results from \textit{w/o RL} indicate that the RL step provides a comprehensive improvement to the model, with progress observed across all datasets. The \textit{SFT+Full} results show that simply increasing the SFT training data does not effectively enhance the model and may even lead to overfitting. Meanwhile, the \textit{SFT+PPO} results demonstrate that GRPO has certain advantages over the previous PPO, offering both memory savings and more effective group-level optimization. The average F1 score of \textit{RAAR-Llama} is comparable to that of RAAR-8b, demonstrating the generality of the RAAR framework.

\section{Error Analysis}

Appendix~\ref{app:erroranalysis} presents detailed confusion metrics, case studies, and reasoning comparisons across \textit{Reason\_zs}, RAAR-8B, and RAAR-14B. The analysis shows that RAAR-14B relies heavily on lexical, stylistic, and sentiment cues. In the Fake News domain, sensational celebrity reports are often flagged as fake, while conventionally styled news is sometimes accepted as legit. In the Rumor dataset, confident or excited unverified claims are frequently classified as non-rumors, whereas humorous or sarcastic posts are misinterpreted as rumors. In the Conspiracy dataset, explicit conspiratorial terminology biases the model toward the \emph{Conspiracy} label, while neutral or technical framings are occasionally under-labeled. Overall, even when the summary agent is prompted for more perspective analysis in addition to the analysis from subagents, RAAR may focus on specific analytical angles in certain cases, potentially reinforcing human-like biases and misinterpreting formally neutral or emotionally restrained content.

The reasoning comparison cases across models show that model scale and inference design critically determine sensitivity to implicit intent and discourse structure, with smaller models relying on surface heuristics and larger models exhibiting deeper, more robust reasoning. More detail analysis can be found at Appendix~\ref{app:erroranalysis}.

\section{Conclusion}

In this paper, we propose the RAAR framework, the first retrieval-augmented agentic reasoning approach for cross-domain misinformation detection. 
RAAR intergreates a retrieval-augmented agentic reasoning mechanism with SFT and RL to enable cross-domain knowledge transfer and multi-step reasoning.
We evaluated our trained models RAAR-8B and RAAR-14B, and competitive baselines on challenging domains selected from three datasets. 
The results demonstrate that RAAR outperforms other cross-domain and LLM-based methods, validating the effectiveness of the framework. Additionally, we conducted ablation studies to analyze the contributions of different components and performed error analyses, providing a foundation for future improvements.

In the future, we aim to enhance the RAAR framework by incorporating more perspectives (e.g., topic and stance) and adapting the framework to multilingual and multimodal settings.

\section{Limitation}

Due to limited computational resources, we only trained the 8B and 14B open-source LLMs. Therefore, we did not explore the potential impact of using larger models or different architectures on the performance of cross-domain misinformation detection when applying the RAAR framework.

Although our analysis considers multiple perspectives, including sentiment, semantics, and writing style, there remain many potentially useful features—such as stance, argumentation structure, and topical information—that were not analyzed. Exploring these features will be a focus of our future work.

Since the framework is designed to focus on sentiment, semantics, and writing style, even though the Summary Agent is instructed to consider other aspects in addition to these three, some cases show that the final reasoning primarily concentrates on sentiment, semantics, and style, resulting in reasoning that is not fully comprehensive.

The COCO dataset contains instances that may belong to multiple topics, making it unrealistic to balance the data and leading to data imbalance. This can also result in the retrieval process occasionally selecting irrelevant or noisy samples.




\bibliography{main}

\appendix

\section{Related Work}

\subsection{Cross-domain misinformation detection}
Cross-domain misinformation detection leverages information or data from different domains to identify misinformation in the target domain, making it more consistent with real-world conditions. \cite{comito2023towards} proposed a deep learning–based architecture that can handle cross-domain fake news by generating high-level features across domains.  \cite{10095281} learns cross-domain transferable features by aligning source news and target news using optimal transport (OT) techniques. \cite{10278156} develops a rough-fuzzy graph learning domain adaptation method for fake news detection that models cross-domain uncertainty, refines target region analysis, and preserves sparse structures to capture shared features. \cite{tong2024mmdfnd} design MMDFND, a multi-modal, multi-domain fake news detection model that leverages domain-aware extraction and cross-modal/domain integration to achieve state-of-the-art results on two real-world datasets. \cite{yue2023metaadapt} proposes MetaAdapt, a meta-learning approach for domain-adaptive few-shot misinformation detection that uses limited target examples to guide adaptive knowledge transfer by weighting source tasks based on their similarity to the meta task. \cite{liu2025raemollm} develops a retrieval-augmented LLM framework for cross-domain misinformation detection that uses affective embeddings to retrieve relevant examples as few-shot demonstrations for in-context learning.  \cite{li2025multi} proposes a multi-agent framework that uses expert agents, a question-reflection mechanism, and decision rule optimization to improve analysis quality and cross-domain performance.
However, existing approaches often rely on task-specific fine-tuning or prompts, consider data from a single perspective without integrating semantics, sentiment, or style, and lack systematic reasoning over ambiguous or implicit evidence. These limitations hinder cross-task and cross-domain generalization, motivating methods that combine robust, interpretable representations with cross-domain reasoning.

\subsection{Application of complex reasoning}

Complex reasoning, represented by GPT-o1 and DeepSeek-R1 \cite{guo2025deepseek}, has further enhanced LLMs and has been applied across multiple fields. Huotuo-o1 \cite{chen2025towards} guides LLMs to construct complex reasoning paths using multi-step strategies and further enhances the model's reasoning ability through supervised fine-tuning and reinforcement learning. \cite{qian2025fino1} introduced FinCoT, the first high-fidelity open CoT corpus for the financial domain, built from seven QA datasets with domain supervision, iterative LLM refinement, and difficulty-aware filtering. Based on FinCoT, they developed the first open financial reasoning model trained with SFT and GRPO. \cite{dai2025psyche} propose Psyche-R1, the first Chinese psychological LLM integrating empathy, expertise, and reasoning, trained on a large corpus of CoT-based questions and empathetic dialogues using a hybrid strategy of GRPO for reasoning and supervised fine-tuning for empathy and domain knowledge. \cite{zhang2025fact} applied a similar process for video misinformation detection. However, these methods primarily rely on direct inference from data within the same distribution, operate with a single model, and lack the capacity to analyze tasks from multiple perspectives, rendering them clearly insufficient for addressing complex cross-domain misinformation.


\begin{algorithm*} 
\footnotesize
	\caption{Retrieval Augmented Agentic Reasoning Training} 
	\label{alg1} 
	\begin{algorithmic}[1]
		\REQUIRE Source data: $D_s$. Target data: $D_t$.  $\{D_s, D_t\} \in D = \{(x, y^*)\}$, SubAgents: $A_{sen}$, $A_{sem}$, $A_{sty}$, Summary Agent: $A_{sum}$, max steps: $N$, Embeddings: $E_{sen}$, $E_{sem}$, $E_{sty}$, Initial model $\pi_\theta$, SubAgents update strategies: double-check $S_{dc}^{sub}$, communication $S_{com}^{sub}$, $S_{random}^{sub}$ denotes randomly select $S_{dc}^{sub}$ or $S_{com}^{sub}$. Summary agent random strategies $S_{random}^{sum} \in \{Refine, Backtracking, NewPath, Validation, Correction\}$, LLM (DeepSeek-V3.2) for searching reasoning paths and a Verifier.
        
		
    \textcolor{blue}{// Stage 1:  Retrieval Augmented Data Building}
    
        \FOR{$e_{sen}^t, e_{sem}^t, e_{sty}^t $ in $E_{sen}(D_t), E_{sem}(D_t), E_{sty}(D_t)$}
            \FOR{$e_{sen}^s, e_{sem}^s, e_{sty}^s $ in $E_{sen}(D_s), E_{sem}(D_s), E_{sty}(D_s)$}
            \STATE $Sco \gets score_{sen}, score_{sem}, score_{sty} = cosine(e_{sen}^t,e_{sen}^s),  cosine(e_{sem}^t,e_{sem}^s), cosine(e_{sty}^t,e_{sty}^s)$
            \ENDFOR
        \ENDFOR
        \STATE $D_{ra} = \{(x_{ra}, y^*)\} \gets$ select top k examples for each dimensions according to $Sco$
        \STATE $D_{search}, D_{RL} \gets split(D_{ra})$ 
        \STATE $D_{SFT} \gets \emptyset$
    
    \textcolor{blue}{// Stage 2: Multi-agent Collaborated Reasoning Path Building}
        \STATE $CoT^{sub}, y^{sub} = SubAgents(D_{search})$
        \STATE $CoT_{0}, y_{0} = A_{sum}(CoT^{sub}, y^{sub})$
        \FOR{$i$ to $N$}
            \IF{$not$ $Verifier(y_{0},y^*)$}
                \IF{$same(y_{sub}) = y_{0}$}
                    \STATE $CoT^{sub}, y^{sub} = SubAgents(CoT^{sub}, y^{sub},S_{dc}^{sub})$
                    \IF{$same(y_{sub})$ and $y_{sub} \neq y_{*}$}
                        \STATE $CoT_{i}, y_{i} = A_{sum}(CoT^{sub}, y^{sub}, S_{random}^{sum})$
                    \ELSE 
                        \STATE $CoT_{i}, y_{i} = A_{sum}(CoT^{sub}, y^{sub})$
                    \ENDIF 
                \ELSE 
                    \STATE $CoT_{i}, y_{i} = A_{sum}(CoT^{sub}, y^{sub}, S_{random}^{sum})$
                    \IF{$not$ $Verifier(y_{i},y^*)$}
                            \STATE $CoT^{sub}, y^{sub} = SubAgents(CoT^{sub}, y^{sub},S_{random}^{sub})$
                            \STATE $CoT_{i}, y_{i} = A_{sum}(CoT^{sub}, y^{sub}, S_{random}^{sum})$
                        \ENDIF 
                \ENDIF 
            \ENDIF
            \IF{$Verifier(y_{0},y^*)$}
                \STATE $\hat{CoT} \gets LLM{rephrase}(CoT_0, CoT_1,...CoT_N,x_{ra})$
                \STATE $\hat{y}  \gets LLM{refine}(\hat{CoT},x_{ra})$
                \STATE $D_{SFT} \gets {(x_{ra}, \hat{CoT}, \hat{y})}$
            \ENDIF
        \ENDFOR

    \textcolor{blue}{// Step 3.1: SFT}
        \FOR{$(x_{ra}, \hat{CoT}, \hat{y})$ in $D_{SFT}$}
            \STATE $L_{SFT}(\theta) \gets -log \pi_{\theta}(\hat{CoT}, \hat{y} \mid x_{ra})$
            \STATE $\theta \gets \theta -\eta \nabla(\theta)L_{SFT}(\theta)$
        \ENDFOR
        
    \textcolor{blue}{// Step 3.2: RL}
        \FOR{$(x_{ra}, y^*)$ in $D_{RL}$}
            \STATE $\hat{CoT}, \hat{y} = \pi_{\theta}(x_{ra})$ 
            \STATE $Reward \gets Rule(\hat{y},y^*) + Format(\hat{CoT}, \hat{y})$
            \STATE $\theta \gets update(L_{GRPO}(x_{ra}, y^*, \hat{CoT}, \hat{y}, Reward, \pi_{\theta})$
        \ENDFOR
    \STATE Return $\pi_{\theta}$
	\end{algorithmic} 
    
\label{alg:algorithm}
\end{algorithm*}

\section{Data analysis \label{app:dataanalysis}}

\subsection{Data process and statistics \label{app:dataprocessstatistics}}

We conduct experiments on the AMTCele \cite{perez2018automatic1}, PHEME \cite{kochkina2018all2}, COCO \cite{langguth2023coco3}. AMTCele was constructed by \cite{liu2025raemollm}, including FakeNewsAMT and Celebrity. It is a cross-domain fake news dataset covering multiple topical domains. Legitimate news is collected from mainstream media outlets, while fake news is generated via Amazon Mechanical Turk crowdsourcing; the Celebrity subset focuses on news from online entertainment and celebrity magazines. PHEME is a Twitter-based dataset consisting of rumours and non-rumours associated with nine breaking news events, widely used to study misinformation detection in social media contexts. COCO is a conspiracy-oriented Twitter dataset covering 12 conspiracy theory categories. Tweets are labeled by overall intention as Unrelated (unrelated to Conspiracy), Related (related to conspiracy but does not propagate), or Conspiracy (related to conspiracy and aims to spread).  

We evaluate the difficulty of each domain in AMTCele and PHEME using GPT-4.1 and DeepSeek-Chat, in order to select relatively challenging domains. The detailed results are reported in Table \ref{tab:difficulty_f1}. For AMTCele, the \textit{celebrity} domain is collected from popular culture media with relatively weak structure, strong contextual dependence, and high semantic noise. As a result, LLMs exhibit comparatively poorer performance on this domain, and we therefore select it for our experiments. For PHEME, based on the performance of the two models and the data volume, we select four more challenging domains: \textit{prince, putinmissing, gurlitt} and \textit{ebola}. Since some of the remaining domains are highly imbalanced (e.g. in charliehebdo domain, 1621 non-rumour and 458 rumour), we apply data balancing for each selected domain, ensuring that the two classes within each domain have approximately equal sample sizes. For COCO, as each instance may involve multiple domains, domain-wise splitting is more complex. Therefore, building upon the RAEmoLLM work, we additionally include data from the Unrelated category to better align with the original task setting of the dataset. The statistics of each dataset can be found in Table \ref{tab:datastatistics}.

\subsection{Sentiment, semantic, and style analysis\label{app:sssanalysis}}

\begin{table*}[]
\footnotesize
\resizebox{1\textwidth}{!}{
\begin{tabular}{lccccccccc}
\hline
AMTCele    & biz         & edu            & entmt        & polit    & sports      & tech        & celebrity    &                      &                      \\ \hline
DeepSeek-C & 0.887       & 0.925          & 0.925        & 0.937    & 0.937       & 0.962       & 0.845        &                      &                      \\
gpt-4.1    & 0.937       & 0.925          & 0.899        & 0.937    & 0.975       & 0.975       & 0.833        &                      &                      \\
Ave.       & 0.912       & 0.925          & 0.912        & 0.937    & 0.956       & 0.969       & {\ul 0.839}  & \multicolumn{1}{l}{} & \multicolumn{1}{l}{} \\ \hline
PHEME      & sydneysiege & ottawashooting & charliehebdo & ferguson & germanwings & prince      & putinmissing & gurlitt              & ebola                \\ \hline
DeepSeek-C & 0.600       & 0.519          & 0.648        & 0.552    & 0.572       & 0.076       & 0.609        & 0.393                & 0.176                \\
gpt-4.1    & 0.652       & 0.563          & 0.626        & 0.602    & 0.593       & 0.394       & 0.385        & 0.343                & 0.300                \\
Ave.       & 0.626       & 0.541          & 0.637        & 0.577    & 0.583       & {\ul 0.235} & {\ul 0.497}  & {\ul 0.368}          & {\ul 0.238}          \\ \hline
\end{tabular}
}
\caption{F1 score of gpt-4.1 and DeepSeek-C in each domain of AMTCele and PHEME. \label{tab:difficulty_f1}}
\end{table*}

\begin{table*}[]
\footnotesize
\resizebox{1\textwidth}{!}{
\begin{tabular}{lllccccccclc}
\hline
        &  & \multicolumn{8}{c}{Train}                                                                                           &  & Test        \\ \cline{1-1} \cline{3-10} \cline{12-12} 
AMTCele &  & Domains     & Technology    & Education    & Business & Sports          & Politics          & Entertainment & Total &  & Celebrities \\ \cline{3-12} 
        &  & Legit       & 40            & 40           & 40       & 40              & 40                & 40            & 240   &  & 250         \\
        &  & Fake        & 40            & 40           & 40       & 40              & 40                & 40            & 240   &  & 250         \\ \hline
PHEME   &  & Events      & Charlie Hebdo & Sydney siege & Ferguson & Ottawa shooting & Germanwings-crash &               & Total &  & 4 events    \\ \cline{3-12} 
        &  & Rumours     & 458           & 522          & 284      & 470             & 238               & -             & 1972  &  & 430         \\
        &  & Non-rumours & 500           & 699          & 300      & 420             & 231               & -             & 2150  &  & 193         \\ \hline
        &  &             & \multicolumn{6}{c}{Fake Virus, Harmful Radiation, Depopulation, and   Unrelated}              & Total &  & Test        \\ \cline{3-12} 
COCO    &  & Unrelated   & \multicolumn{6}{c}{604}                                                                       & 604   &  & 302         \\
        &  & Related     & \multicolumn{6}{c}{540}                                                                       & 540   &  & 248         \\
        &  & Conspiracy  & \multicolumn{6}{c}{1181}                                                                      & 1181  &  & 612         \\ \hline
\end{tabular}
}
\caption{Statistics of datasets. AMTCele includes 7 domains. PHEME contains 9 domains (events). COCO has 12 domains (topics) and Unrelated category. For AMTCele, we apply Celebrity as the test set, and for PHEME, we apply four difficult domains as the test set. For COCO, we select 3 domains and part of Unrelated data as the test set. \label{tab:datastatistics}}
\end{table*}

\begin{table*}[]
\footnotesize
\resizebox{1\textwidth}{!}{
\begin{tabular}{llcccccccccccccc}
\hline
AMTCele     &  & \multicolumn{4}{c}{Semantic (Roberta)}                        &           & \multicolumn{4}{c}{Sentiment (EmoLlama)}                      &           & \multicolumn{4}{c}{Writing style   (styleembedding)}          \\ \cline{3-6} \cline{8-11} \cline{13-16} 
            &  & \textbf{top1} & \textbf{top2} & \textbf{top4} & \textbf{top8} & \textbf{} & \textbf{top1} & \textbf{top2} & \textbf{top4} & \textbf{top8} & \textbf{} & \textbf{top1} & \textbf{top2} & \textbf{top4} & \textbf{top8} \\ \hline
fake→fake   &  & 0.99          & 0.99          & 0.99          & 0.99          &           & 0.86          & 0.85          & 0.84          & 0.82          &           & 0.74          & 0.73          & 0.73          & 0.72          \\
fake→legit  &  & 0.99          & 0.99          & 0.99          & 0.99          &           & 0.81          & 0.80          & 0.77          & 0.75          &           & 0.73          & 0.72          & 0.71          & 0.70          \\
t           &  & 3.96          & 5.03          & 6.37          & 8.77          &           & 4.91          & 7.42          & 11.44         & 17.97         &           & 5.45          & 8.01          & 11.46         & 16.59         \\
p           &  & 0.00          & 0.00          & 0.00          & 0.00          &           & 0.00          & 0.00          & 0.00          & 0.00          &           & 0.00          & 0.00          & 0.00          & 0.00          \\
legit→legit &  & 0.99          & 0.99          & 0.99          & 0.99          &           & 0.89          & 0.88          & 0.87          & 0.85          &           & 0.73          & 0.73          & 0.72          & 0.71          \\
legit→fake  &  & 0.99          & 0.99          & 0.99          & 0.99          &           & 0.87          & 0.86          & 0.84          & 0.83          &           & 0.73          & 0.72          & 0.71          & 0.70          \\
t           &  & 2.78          & 4.64          & 7.59          & 11.31         &           & 2.30          & 3.49          & 5.42          & 9.18          &           & 3.14          & 3.94          & 4.97          & 6.12          \\
p           &  & 0.01          & 0.00          & 0.00          & 0.00          &           & 0.02          & 0.00          & 0.00          & 0.00          &           & 0.00          & 0.00          & 0.00          & 0.00          \\ \hline
\end{tabular}
}
\caption{Statistics values of cosine similarity between embeddings of different dimensions on AMTCele dataset. Top K denotes retrieval top K examples. “A→B” represents the calculation of cosine similarity between each data point in A category in one domain and each data point in B in other domains. Each element (i, j) in the resulting calculation represents the cosine similarity between the i-th vector in the A group embeddings and the j-th vector in the B group embeddings. The top 4 refers to selecting the four highest values from each row. The t-value and p-value represent the t-test results for the “A-B” results of the two lines above. \label{tab:valuessimilarity}}
\end{table*}

\begin{table*}[]
\footnotesize
\resizebox{1\textwidth}{!}{
\begin{tabular}{lccccccccccccccccccc}
\hline
AMTCele     & \multicolumn{4}{c}{all-mpnet-base-v2}                                                                                                         &                               & \multicolumn{4}{c}{Qwen3-Embedding-8B}                                                                                                        &                               & \multicolumn{4}{c}{StyleDistance}                                                                                                             &                               & \multicolumn{4}{c}{Sentibert}                                                                                                                 \\ \cline{2-5} \cline{7-10} \cline{12-15} \cline{17-20} 
\textbf{}   & \multicolumn{1}{l}{\textbf{top1}} & \multicolumn{1}{l}{\textbf{top2}} & \multicolumn{1}{l}{\textbf{top4}} & \multicolumn{1}{l}{\textbf{top8}} & \multicolumn{1}{l}{\textbf{}} & \multicolumn{1}{l}{\textbf{top1}} & \multicolumn{1}{l}{\textbf{top2}} & \multicolumn{1}{l}{\textbf{top4}} & \multicolumn{1}{l}{\textbf{top8}} & \multicolumn{1}{l}{\textbf{}} & \multicolumn{1}{l}{\textbf{top1}} & \multicolumn{1}{l}{\textbf{top2}} & \multicolumn{1}{l}{\textbf{top4}} & \multicolumn{1}{l}{\textbf{top8}} & \multicolumn{1}{l}{\textbf{}} & \multicolumn{1}{l}{\textbf{top1}} & \multicolumn{1}{l}{\textbf{top2}} & \multicolumn{1}{l}{\textbf{top4}} & \multicolumn{1}{l}{\textbf{top8}} \\ \hline
fake→fake   & 0.41                              & 0.38                              & 0.36                              & 0.33                              &                               & 0.44                              & 0.42                              & 0.40                              & 0.38                              &                               & 0.93                              & 0.93                              & 0.93                              & 0.92                              &                               & 0.85                              & 0.84                              & 0.83                              & 0.82                              \\
fake→legit  & 0.40                              & 0.38                              & 0.35                              & 0.32                              &                               & 0.42                              & 0.41                              & 0.39                              & 0.37                              &                               & 0.92                              & 0.92                              & 0.91                              & 0.91                              &                               & 0.85                              & 0.84                              & 0.83                              & 0.82                              \\
t           & 0.79                              & 1.39                              & 2.41                              & 3.98                              &                               & 4.51                              & 6.61                              & 9.85                              & 14.31                             &                               & 3.90                              & 6.05                              & 9.15                              & 14.45                             &                               & 2.15                              & 2.33                              & 2.28                              & 2.08                              \\
p           & 0.43                              & 0.17                              & 0.02                              & 0.00                              &                               & 0.00                              & 0.00                              & 0.00                              & 0.00                              &                               & 0.00                              & 0.00                              & 0.00                              & 0.00                              &                               & 0.03                              & 0.02                              & 0.02                              & 0.04                              \\
legit→legit & 0.41                              & 0.38                              & 0.35                              & 0.32                              &                               & 0.42                              & 0.40                              & 0.38                              & 0.36                              &                               & 0.95                              & 0.95                              & 0.94                              & 0.94                              &                               & 0.85                              & 0.84                              & 0.84                              & 0.82                              \\
legit→fake  & 0.40                              & 0.38                              & 0.35                              & 0.32                              &                               & 0.42                              & 0.41                              & 0.39                              & 0.37                              &                               & 0.93                              & 0.92                              & 0.92                              & 0.91                              &                               & 0.84                              & 0.84                              & 0.83                              & 0.82                              \\
t           & 1.36                              & 1.31                              & 1.36                              & 0.92                              &                               & -0.25                             & -1.25                             & -3.18                             & -5.85                             &                               & 9.21                              & 13.35                             & 19.37                             & 27.75                             &                               & 3.16                              & 4.47                              & 5.67                              & 6.84                              \\
p           & 0.17                              & 0.19                              & 0.17                              & 0.36                              &                               & 0.81                              & 0.21                              & 0.00                              & 0.00                              &                               & 0.00                              & 0.00                              & 0.00                              & 0.00                              &                               & 0.00                              & 0.00                              & 0.00                              & 0.00                              \\ \hline
\end{tabular}
}
\caption{Statistics values of cosine similarity from other embedding models. Top K denotes retrieval top K examples. “A→B” represents the calculation of cosine similarity between each data point in A category in one domain and each data point in B in other domains. Each element (i, j) in the resulting calculation represents the cosine similarity between the i-th vector in the A group embeddings and the j-th vector in the B group embeddings. The top 4 refers to selecting the four highest values from each row. The t-value and p-value represent the t-test results for the “A-B” results of the two lines above. \label{tab:valuessimilarity_others}}
\end{table*}

We take AMTCele as an example for analysis, since the number of legit and fake data points is equal in each domain, making the analysis results more reasonable and reliable. Table \ref{tab:valuessimilarity} presents the statistics of similarities for different embeddings (semantic \cite{liu2019roberta}, sentiment \cite{liu2024emollms}, and writing style \cite{wegmann2022same}). We conducted t-tests on the top K cosine similarities both within and between categories. For example, “fake-legit” refers to the cosine similarity computed between each data point in the “fake” category and each data point in the “legit” category. In this computation, the “fake” data points come from one domain, while the “legit” data points come from other domains. The results shown are the averages across all domains. We then selected the top K similarity values and performed t-tests on them. The results indicate that the t-values for the top K (1, 2, 4, 8) similarity values, both within and between categories, are all positive, with p-values less than 0.05. This suggests that the top K data points retrieved based on cosine similarity within the “fake” category are highly likely to mostly belong to the same “fake” category. From the results in Table \ref{tab:valuessimilarity}, it can be seen that all embeddings lead to the same conclusion for the top 1, 2, 4, and 8 cases. Considering the ablation study presented in the main text, including the 1-shot and 4-shot settings, we ultimately adopt the 2-shot approach used in this paper, retrieving 2 examples for the experiments. Additionally, we also conduct similar analysis based on other kinds of semantic embeddings (all-mpnet-base-v2 \cite{song2020mpnet}, and Qwen3-Embedding-8B \cite{qwen3embedding}), sentiment embeddings (sentibert \cite{yin2020sentibert}), and style embeddings (StyleDistance \cite{patel2025styledistance}). Table \ref{tab:valuessimilarity_others} presents their statistical results. Since their p-values are higher than those of the aforementioned models and may result in greater bias, we adopt the three models described above.

\section{Experiment settings and baseline details}

\subsection{Experiment settings \label{app:settings}}

We develop two LLMs based on the RAAR framework, namely RAAR-8B and RAAR-14B, which are built upon the reasoning versions of Qwen3-8B and Qwen3-14B, respectively. In stage 1, we select the top two examples for each dimension, details are provided in Section \ref{sec:ablation}. Both models are trained in two stages, consisting of SFT followed by RL optimization. During SFT, we set the learning rate to 1e-5 with a warmup ratio of 0.1. Training is conducted for 3 epochs with a batch size of 128. In the RL stage, we perform 8 rollouts per sample with a training batch size of 1024. The model is trained for 5 epochs using a learning rate of 1e-6 and a sampling temperature of 1. We adopt DeepSpeed ZeRO-3 optimization and offload parameters to the CPU to further save GPU memory. For both stages, the maximum input sequence length is 24k tokens, and the maximum output length is 8k tokens. Full-parameter training is carried out on four NVIDIA A100 GPUs (80 GB) in both stages.

\subsection{Baseline details \label{app:baselines}}

\textbf{Cross-domain methods:} MOSE \cite{qin2020multitask} adopts a multi-domain mixture-of-experts (MoE) architecture with a dedicated prediction head for each domain. EDDFN \cite{silva2021embracing} is designed to jointly capture domain-specific features and domain-invariant representations. MDFEND \cite{nan2021mdfend} introduces a Domain Gate mechanism to dynamically select the most relevant experts within an MoE framework. CANMD \cite{yue2022contrastive} addresses label distribution shift through label shift correction and leverages contrastive learning to enhance domain generalization. MetaAdapt \cite{yue2023metaadapt} employs a meta-learning paradigm to enable domain-adaptive few-shot misinformation detection.

\textbf{LLM-based Adaptation methods:} \textbf{Instruction Tuning}: We fine-tune two variants of the model based on Qwen3-8b, including a zero-shot fine-tuned version (IT\_zs) (like conspemollm \cite{liu2024conspemollm}) and a retrieval-augmented fine-tuned version with few-shot examples (IT\_ra). \textbf{RAEmoLLM} \cite{liu2025raemollm} is a misinformation detection framework that enhances in-context demonstration retrieval with emotional information; in our implementation, GPT-4.1 is adopted as the base model, and examples are retrieved based on sentiment intensity. \textbf{MARO} \cite{li2025multi} is a multi-agent framework that improves cross-domain misinformation detection by integrating expert agents, question-driven reflection, and cross-domain decision rule optimization.

\subsection{Prompt templates for evaluating LLMs \label{app:template4LLMs}}

The following are the templates for zero-shot inference and few-shot inference based on RA data. \textit{[task description]} for AMTCele: \textit{``Determine whether the target text is fake or legit.''}. \textit{[task description]} for PHEME: \textit{``Classify the target text as rumour or non-rumour.''}.  \textit{[task description]} for COCO: \textit{``Determine whether the target text is Unrelated, Related, or Conspiracy. Unrelated means the text contains conspiracy-related keywords but uses them in an unrelated or different context. Related means the text is conspiracy-related but does not propagate conspiracy misinformation. Conspiracy means the text is conspiracy-related and actively propagates or supports the misinformation.''}. \textit{[target text]} and  is the original text from each dataset. The zero-shot and few-shot instruction-tuning methods are also based on these two templates. 

\begin{center}
\footnotesize
\fcolorbox{black}{gray!10}{
\begin{minipage}{0.45\textwidth}
\footnotesize
\textbf{Template for constructing SFT dataset}  \\
\textbf{Input:} $x_{ra}$\\
\textbf{Output:} \\
<think>\\
\textit{[$\hat{CoT}$]} \\
</think> \\

<answer> \\
\textit{[$\hat{y}$]} \\
</answer>

\end{minipage}
}
\end{center}

\begin{center}
\footnotesize
\fcolorbox{black}{gray!10}{
\begin{minipage}{0.45\textwidth}
\footnotesize
\textbf{Template for Zero-shot setting}  \\
\textbf{Task:} \textit{[task description]}\\
\textbf{Target text:} \textit{[target text]}
\end{minipage}
}
\end{center}

\begin{center}
\footnotesize
\fcolorbox{black}{gray!10}{
\begin{minipage}{0.45\textwidth}
\footnotesize
\textbf{Template for few-shot setting (same with $x_{ra}$})  \\
\textbf{Task:} \textit{[task description]} \\
\textbf{Target text:} \textit{[target text]} \\

Here are a few examples retrieved through sentiment intensity: \\
\textbf{Text1}: \textit{[text1]} \\
\textbf{Text2}: \textit{[text2]} \\

Here are a few examples retrieved through semantic information: \\
\textbf{Text1}: \textit{[text1]} \\
\textbf{Text2}: \textit{[text2]} \\

Here are a few examples retrieved through writing style information: \\
\textbf{Text1}: \textit{[text1]} \\
\textbf{Text2}: \textit{[text2]} \\

\end{minipage}
}
\end{center}

\begin{figure*}[t]  
\centering

\begin{subfigure} 
\centering
\includegraphics[width=1.5\columnwidth]{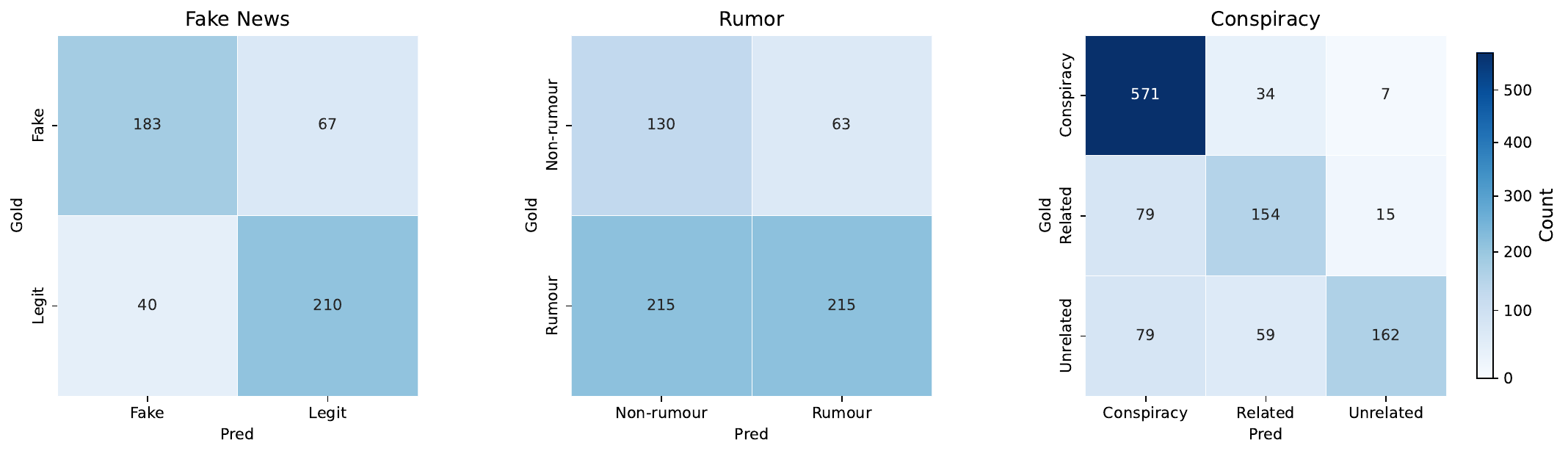}
\caption{Confusion Metrics of zero-shot reasoning models (Reason\_zs ablation).}
\label{fig:Confusion_8b_zsreasoning}
\end{subfigure}

\vspace{0.5em} 

\begin{subfigure}
\centering
\includegraphics[width=1.5\columnwidth]{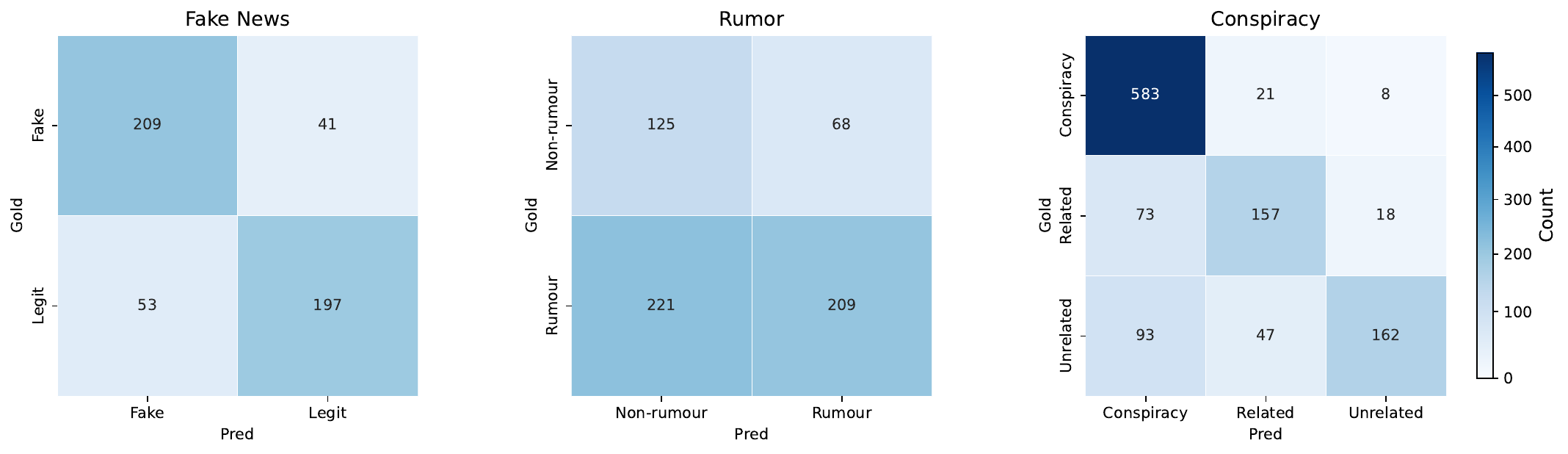}
\caption{Confusion Metrics of RAAR-8b.}
\label{fig:Confusion_8b_RAAR}
\end{subfigure}

\vspace{0.5em}

\begin{subfigure}
\centering
\includegraphics[width=1.5\columnwidth]{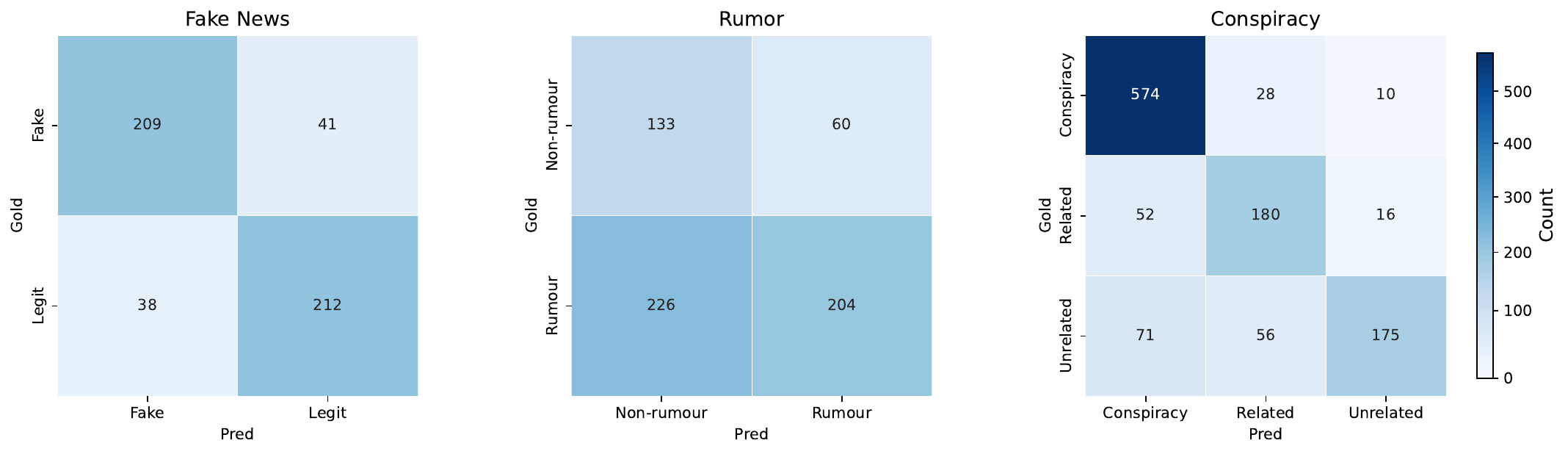}
\caption{Confusion Metrics of RAAR-14b.}
\label{fig:Confusion_14b_RAAR}
\end{subfigure}

\end{figure*}

\section{Error Analysis Details \label{app:erroranalysis}}

\subsection{Confusion Metrics}

\subsection{Case Study for RAAR-14b \label{app:case_14b}}

In this section, we present a case study of RAAR-14b by closely examining its predictions, as it achieves the strongest performance across all evaluation settings. To save space, we have not shown retrieval examples for all instances.

\paragraph{Case Study 1}

\textbf{Dataset:} AMTCele

\textbf{Target Text:}
\begin{quote}\small
Jennifer Aniston Finally Pregnant At 48 Years Old? | Celeb Dirty Laundry

Jennifer Aniston and Justin Theroux have been married more than a year, so the inevitable pregnancy rumors will begin. According to the latest cover issue of Star Magazine, Jennifer Aniston is ``finally'' pregnant at 48 years old. Now, we've heard in the past that Jennifer Aniston has struggled to get pregnant, even back when she was married to Brad Pitt.
\end{quote}

\textbf{Error Type:} Legit $\rightarrow$ Fake

\textbf{Analysis:}
This article was labeled as fake primarily due to its sensational headline and tabloid style framing. Although the content reports on a celebrity rumor circulating in mainstream entertainment media, the model appears to associate speculative language and celebrity topics with falsity. The presence of a question mark in the headline and emotionally charged phrasing likely triggered a fake prediction, despite the text functioning as standard entertainment reporting rather than fabricated misinformation. This case illustrates how stylistic sensationalism is treated as a proxy for falsity in the model’s decision process.

\paragraph{Case Study 2}

\textbf{Dataset:} AMTCele

\textbf{Target Text:}
\begin{quote}\small
Brad Pitt Is ``Still in Contact'' With Ex Jennifer Aniston Following Angelina Jolie Divorce (REPORT)

According to a new report by E! Online, Brad Pitt is still in communication with his ex-wife, Jennifer Aniston. A source tells the outlet that the pair have a ``friendly, but limited relationship.'' The confirmation comes shortly after Us Weekly reported that Pitt had been texting Aniston amid his divorce from Angelina Jolie.
\end{quote}

\textbf{Error Type:} Fake $\rightarrow$ Legit

\textbf{Analysis:}
This fabricated gossip article was misclassified as legitimate because it closely follows conventional news-reporting conventions. It cites recognizable media outlets, attributes claims to unnamed sources, and adopts a neutral narrative tone. These surface signals resemble professional journalism, leading the model to accept the content as legitimate. The error shows that the model relies on reporting format and stylistic realism rather than assessing whether the claims are independently verifiable or speculative in nature.

\paragraph{Case Study 3}

\textbf{Dataset:} PHEME

\textbf{Target Text:}
\begin{quote}\small
TORONTO: Quick reminder that Prince is seemingly playing @masseyhall tonight. No more details at the minute.

Here are a few examples retrieved through sentiment intensity:
 Text: Ottawa, be safe, be smart.. The Hill is under attack. The label of this text: non-rumour.
  Text: We are in full lock down until further notice from Ottawa Police.. The label of this text: rumour.

Here are a few examples retrieved through semantic information:
 Text: The \#sydneysiege is over! Confirmed by Sydney Police on twitter!
Thank god! I hope everyone are and stays safe!. The label of this text: non-rumour.
  Text: Ottawa police are confirming a shooting at the War Memorial. Minutes ago. No other info. \#cbcOTT \#OTTnews. The label of this text: non-rumour.

Here are a few examples retrieved through writing style information:
 Text: BREAKING: Gunman in \#SydneySiege has told police he has "devices all over the city" \&amp; demands to speak with Prime Minister, acc. to reports.. The label of this text: rumour.
  Text: The door is solid, you can’t knock it down… it’s bulletproof-\@JimTilmon on \@NYTimes rpt \#Germanwings pilot locked out. The label of this text: rumour.

\end{quote}

\textbf{Error Type:} Rumour $\rightarrow$ Non-rumour

\textbf{Analysis:}
This post conveys unverified information about a possible surprise performance, indicated by vague phrasing such as “seemingly” and the absence of official confirmation. However, the model classified it as non-rumour, likely because the message is presented confidently and resembles an announcement rather than speculation. The lack of explicit uncertainty markers, such as “might” or “possibly,” led the model to treat the statement as factual. Compared to few-shot examples, the model often relies on overt cues: posts labeled as rumours frequently contain urgent language, emotional intensity, or dramatic stylistic features (e.g., “BREAKING,” “all over the city”), whereas this post’s neutral, matter-of-fact style aligns more closely with non-rumour examples that report confirmed events. This demonstrates that the model struggles to recognize implicit uncertainty in short social media posts and tends to be misled when unverified information is framed confidently or in a formal announcement style.
\paragraph{Case Study 4}

\textbf{Dataset:} PHEME

\textbf{Target Text:}
\begin{quote}\small
\#putindead I heard it from a good source that Vladimir Putin was riding a weasel on the back of a woodpecker and he fell off and died.
\end{quote}

\textbf{Error Type:} Non-rumour $\rightarrow$ Rumour

\textbf{Analysis:}
This tweet is clearly satirical, using absurd imagery to mock death rumors. Despite this, the model labeled it as a rumour, likely due to literal interpretation of phrases such as ``heard it from a good source'' and the mention of a public figure’s death. The model fails to detect humor and exaggeration, treating the post as a serious information claim rather than parody. This reflects an over-reliance on lexical cues without pragmatic interpretation.






\paragraph{Case Study 5}

\textbf{Dataset:} COCO

\textbf{Target Text:}
\begin{quote}\small
Good luck \& Good luck; remember until Trump gets COVID it's just a Democrat-Deep State-Pedophile Hoax to undermine Trump.  BTW Jared me; a Nigerian prince have come up with a vaccine based on experiments by China on Uigher prisoners. Plz invest in our company Total Fraud Inc.

Here are a few examples retrieved through sentiment intensity:
 Text: That's why they're called the Lying Fake News and they can cover for the Deep State Democrats and their Chinese friends we were not supposed to find out about Flynn Trump or the Wuhan Virus origin so they could monopolize the drug and vaccine market and America gets the short end. The label of this text: Conspiracy (related and supporting).
  Text: Love how the Q people are willing to believe Soros pays protesters; Hillary eats babies but can't accept that Trump got COVID from hosting superspreader events; rallies; making fun of people with masks on...they think it was a Deep State Op or something. The label of this text: Related (but not supporting).

Here are a few examples retrieved through semantic information:
 Text: We'd better isolate the Trump cult so only they can be hurt by their ignorance; indifference to science.    I'm worried they'll stop Soc Dist; spread Covid-19. I hear them saying it's a "deep state lib conspiracy" to hurt Trump. The label of this text: Related (but not supporting).
  Text: Pfizer invested and two Muslim scientists Dr. Şahin and Türeci invented the vaccine for COVID-19 and Trump did what? Shamelessly invented a lie. Maybe his Muslim ban was his input. Trumptards wonder if vaccine is a deep state hoax that they can trust. Ignorance is bliss until.... The label of this text: Related (but not supporting).

Here are a few examples retrieved through writing style information:
 Text: Even Trump said in a recent presser "we're doing this [insert Deep State/Drain the Swamp with COVID19 shutdown] for them...the children. Present and future generations.  I'm a Healthcare care worker  forced to take PTO (layoff) because of fear.. The label of this text: Related (but not supporting).
  Text: Local school gets closed for a week due to a Covid outbreak:  Cue local moms screaming that immuno-compromised people should stay home; referring to this as a "Plandemic".  This is why case numbers are going up Folks. Selfishness and the death of compassion and common sense.. The label of this text: Unrelated.

\end{quote}

\textbf{Error Type:} Conspiracy $\rightarrow$ Unrelated

\textbf{Analysis:}
This post explicitly advances a conspiratorial framing by labeling COVID as a hoax and attributing it to a coordinated plot involving Democrats, the Deep State, and pedophiles. Despite these strong conspiracy indicators, the model predicted Unrelated, suggesting it failed to recognize the conspiratorial claim as the main communicative function. A likely reason is that the sentence is short, colloquial, and the conspiratorial content is compressed into a single compound phrase. Compared with few-shot examples retrieved through sentiment and semantic information, such as texts explicitly linking Trump, the Deep State, or vaccine hoaxes and labeled Conspiracy (related and supporting), this post presents similar thematic content but in a denser, slogan-like format. If the model relies on a narrow set of keyword triggers or expects more elaborated narrative structures, it may mis-handle compressed conspiracy jargon and dismiss it as off-topic rhetoric. This case highlights the model’s under-sensitivity to conspiracy cues when they appear as condensed, informal expressions rather than extended argumentative text.

\subsection{Case Study for Model Performance Comparison}


Across model variants, performance differences are primarily driven by model scale and inference configuration. Smaller zero-shot models rely heavily on shallow heuristics, over-weighting surface cues such as negation, hedging, and source mentions, which results in systematic errors in cases requiring discourse-level integration, including rumor initiation through questioning, denial-as-rumor dynamics, and implicit conspiracy endorsement. Prompting improves task alignment for smaller models, but their predictions remain dominated by explicit lexical patterns rather than integrated reasoning. In contrast, larger models demonstrate stronger holistic reasoning, jointly integrating sentiment, semantics, and pragmatic context. This enables more robust handling of mixed stances, rhetorical camouflage, and implicit presuppositions, leading to more accurate distinctions between discussion and propagation, skepticism and endorsement, and factual reporting and rumor circulation. Overall, increased model scale enhances sensitivity to implicit intent and discourse structure, whereas smaller models remain prone to reasoning fragmentation when surface cues conflict with communicative function.

\begin{figure*}[t]
\centering
\fcolorbox{black}{gray!10}{
\begin{minipage}{\textwidth}
\tiny
\textbf{Model Reasoning Performance Comparison Case Study 1 - Fake News}
\begin{quote}
\textbf{Dataset Name.} AMTCele

\textbf{Instance Content (Original Text).}

\emph{Jennifer Aniston dashes `Friends' reunion hopes}

Jennifer Aniston has once again broken hearts by shooting down the possibility of a ``Friends'' reunion.
In an interview with ITV, the actress talked about looking back with fondness on working on the beloved, hit NBC series.
``Not only was it a gift for us, but it's something people have been able to carry with them,'' Aniston said. ``It's comfort food and it makes them feel better when they are feeling down, when they want to distract themselves.''
And to answer the question that always seems to come up -- will there be a reunion? -- Aniston said don't count on it.
``I don't know what we would do,'' she said. ``I think that period of time was sort of nostalgic, you know?'' ``There's something about a time where our faces weren't shoved into cell phones and we weren't, like, checking Facebooks and Instagrams, and we were in a room together -- or at a coffee shop together -- and we were talking,'' she added. ``And we've lost that.'' But they can still ``be there for you'' in syndication, so hold tight to that. You can, however, see Aniston on the big screen when her latest film Office Christmas Party opens December~9.

\vspace{0.5em}
\textbf{Gold Label.} Legit

\vspace{0.5em}
\textbf{8B Zero-shot Prediction.} Fake

\textbf{8B Zero-shot Reasoning.}
Okay, let's look at this text about Jennifer Aniston and a `Friends' reunion. Hmm, it's written like a typical celebrity news article—has a headline, quotes her directly, and even mentions a specific interview source, ITV. That feels pretty legit at first glance.

Wait, but I remember that `Friends' reunion rumors have been all over the place for years. I think there was a big one a few years back where they actually did a special episode or something. So if Aniston said there wouldn't be a reunion, that might be a bit surprising or even contradictory to what happened before.

Let me think... Oh right, there was a Friends: The Reunion special on Netflix in 2021. That was a huge deal. So if this article is saying she shot down the idea, but then a reunion actually happened, that's a problem. The text says she said `don't count on it' and that the period was `nostalgic' and `we've lost that.' But then the Netflix special was a full reunion with all the cast. That seems like a direct contradiction.

Also, the article mentions her new movie Office Christmas Party opening December~9. I think that movie came out in 2017. So the timing here feels off. If this were a real article from when the movie was coming out, it would have been around late 2017. But the Netflix reunion special was in early 2021. So if this article was published in late 2017, it would be predicting something that hadn't happened yet. But if it was published after the reunion, it would be wrong because the reunion did happen.

Let me check the quotes again. She says, `I think that period of time was sort of nostalgic' and `we've lost that.' That sounds like she's talking about the past, not ruling out a future event. Saying `don't count on it' is vague—it could mean she doesn't know, or she doesn't want to speculate. But the Netflix reunion was a real event, so her statement here seems to be wrong.

Oh, and the article is presented as a news report, but it's not from a major outlet like The New York Times or Entertainment Weekly. It's just `ITV' as the source. That's a bit suspicious—real news outlets usually attribute quotes to specific journalists or have more context. This feels like it could be a fabricated article, maybe from a fake news site, that's using a real quote from a real interview but then making up the whole story about a `Friends' reunion being shot down, when in fact the opposite happened.

So putting it all together: the core claim—that Aniston said there wouldn't be a `Friends' reunion—is false because a real reunion happened. The article uses real quotes and a real interview source to make it seem credible, but the main point is fabricated. That's classic fake news: take a few real details to make a false story believable. I think this is fake.

\vspace{0.5em}
\textbf{RAAR-8B Prediction.} Legit

\textbf{8B Reasoning.}
Okay, let me start by reading this target text carefully. It's about Jennifer Aniston supposedly killing `Friends' reunion hopes. Hmm, the headline is pretty dramatic - `dashes hopes' and `broken hearts' - that feels like clickbait language right away.

Wait, but let me check the sentiment examples they provided. The legit `Walking Dead' example has emotional language too - `heartbreaking,' `bittersweet' - but it's presented as genuine interview content. The fake Scarlett Johansson example feels different though, with that awkward `Easter Eggs' line that doesn't sound like real conversation. Comparing this to the target text, the Aniston quotes do sound natural and emotional, not forced. She's talking about comfort food and nostalgia, which feels authentic.

Now looking at the semantic content - does this make sense? Friends reunion rumors are totally real, and Aniston has been pretty clear about not doing one. The details about how the cast used to interact before social media... that actually rings true based on what I know about the show's history. And the mention of her new movie Office Christmas Party - I think that was real, released around December~2017. This all fits together logically.

Oh, and the writing style - it reads like proper journalism. It's structured like a real news article with attribution, quotes, and background context. The fake examples tend to have more sensational language or awkward phrasing, but this feels professional. The transition from the emotional interview to the business update about her film seems smooth too.

Let me double-check - is there anything that feels off? The quotes sound like something Aniston would actually say. The interview with ITV seems plausible. The timeline checks out. Nothing jumps out as obviously fabricated like that awkward `Easter Eggs' line in the fake example.

Actually, thinking about it more - this reminds me of how real entertainment news often works. Celebrities do express nostalgia about old shows, and the social media comparison feels like genuine commentary. The article even acknowledges the irony of being on phones now versus being together in person.

I'm leaning toward this being legit. The emotional tone matches real interview content, the facts check out, and the writing style is professional. It doesn't have the red flags I saw in the fake examples. The sentiment feels authentic rather than manufactured for clicks.

\vspace{0.5em}
\textbf{RAAR-14B Prediction.} Legit

\textbf{14B Reasoning.}
Okay, let me start by reading this target text carefully. It's about Jennifer Aniston saying there won't be a Friends reunion. Hmm, the tone feels right - she's nostalgic but realistic, not overly emotional or sensational. That reminds me of the legit examples where actors talk about their work in a measured way.

Wait, let me check the sentiment examples they provided. The fake Scarlett Johansson one felt different - it had this weird mix of reluctance and excitement that didn't quite match. But this Aniston piece? She's genuinely reflective about the past, talking about how the show provided comfort, and then clearly states there's no reunion. The emotional flow feels natural, not manufactured.

Now looking at the semantic details - oh, this is important. The text mentions specific things: ITV interview, December~9 release date for her movie, the exact quote structure with `I think that period of time was sort of nostalgic.' These aren't vague claims - they're concrete details that would be hard to fake convincingly. Legit news usually includes these verifiable elements.

Also, the writing style feels professional. It's structured like real journalism - headline, attribution, direct quotes, background context. Reminds me of the Columbia acceptance error story they labeled as legit. The language is formal but accessible, with proper punctuation and balanced sentences.

Let me double-check against the fake patterns. The fake examples often have inconsistencies or emotional whiplash. But here, everything hangs together logically. She talks about the show's legacy, acknowledges the reunion question, gives her answer, and then naturally transitions to her current project. No red flags in the narrative flow.

Hmm, all three angles - sentiment, semantics, and style - seem to point in the same direction. The sentiment is authentic nostalgia, the details are specific and verifiable, and the writing matches professional standards. There's no internal contradiction that would make me suspicious.

Oh, and thinking about it independently - I remember hearing about Jennifer Aniston discussing Friends reunions in interviews. The content aligns with what I'd expect from a real celebrity interview. Nothing feels exaggerated or implausible.

So putting it all together: the emotional tone feels genuine, the specific details check out, the writing is professional, and it matches real-world knowledge. Everything consistently suggests this is legitimate reporting rather than fabricated content.
\end{quote}

\textbf{Analysis:} This case highlights clear differences in reasoning reliability across model scales and inference paradigms. The 8B zero-shot model misclassifies the instance by over-relying on external world knowledge and post-hoc factual verification (e.g., later real-world reunion events), introducing temporal leakage that is inappropriate for the task setting. In contrast, both the 8B and 14B models correctly predict the label by grounding their decisions in intrinsic textual cues, including narrative coherence, quotation structure, and stylistic consistency, rather than speculative external validation. 

Comparing the two non-zero-shot models, the 14B model demonstrates more stable and concise reasoning, integrating sentiment, semantics, and writing style into a unified judgment without overfitting to individual cues. The 8B model reaches the correct conclusion but relies on more surface-level heuristics and explicit pattern matching. Overall, this example illustrates that larger models exhibit stronger capacity for holistic reasoning and internal evidence integration, while smaller zero-shot models are more prone to reasoning drift when unsupported by task-aligned inductive biases.

\end{minipage}
}
\end{figure*}

\begin{figure*}[t]
\centering
\fcolorbox{black}{gray!10}{
\begin{minipage}{\textwidth}
\tiny

\textbf{Model Reasoning Performance Comparison Case Study 2 - Rumour}

\begin{quote}
\textbf{Dataset Name.} PHEME

\vspace{0.5em}
\textbf{Instance Content (Original Text).}

Security at Massey Hall tells \@CP24 there is NO Prince show \#Toronto \#prince'd \#purple'd

Here are a few examples retrieved through sentiment intensity:
 Text: "Contrary to earlier reports no incident occurred near the Rideau Centre," Ottawa police say. \#cbcOTT \@OTTnews. The label of this text: rumour.
  Text: \#Ottawa City Hall is currently in lock-down. Please avoid the area.. The label of this text: non-rumour.

Here are a few examples retrieved through semantic information:
 Text: Rideau Centre general manager tell \@globeandmail there was no shooting inside the mall. But people can't leave or enter. \#OttawaShooting. The label of this text: rumour.
  Text: \#CNN frustrated by Canadians' lack of panic. \#OttawaShooting \#cdnpoli. The label of this text: non-rumour.

Here are a few examples retrieved through writing style information:
 Text: All my prayers are going to the hostages in Sydney..
And all not Muslims are terrorists..It's a religion of peace.. The label of this text: non-rumour.
  Text: Harper to make statement today after gunman shot dead in \#Parliament. PM safe, not on Parliament Hill, being briefed by security officials.. The label of this text: rumour.

\vspace{0.5em}
\textbf{Gold Label.} Rumour

\vspace{0.5em}
\textbf{8B Zero-shot Prediction.} Non-rumour

\textbf{8B Zero-shot Reasoning.}
Okay, let's look at this text. It says security at Massey Hall told CP24 there's no Prince show. Hmm, first impression—this sounds like a direct statement from an official source, not some random guess.

Wait, but the task is to figure out if it's a rumour or not. What makes something a rumour? Usually, it's unverified information that's spreading, right? Like when people share something without knowing if it's true, and it gets passed around.

Let me think about this specific claim. Prince was a huge artist, and Massey Hall is a big venue in Toronto. I remember there was a lot of buzz about a Prince show there a few years back. So this isn't some obscure event—it's a real, high-profile thing that people were talking about.

Now, the text says security specifically told CP24, which is a local news station. That's a credible source. If security at the venue is saying there's no show, that's pretty definitive. It's not like they're saying 'maybe' or 'I heard something.' It's a flat 'NO.'

Also, the hashtags \#Toronto and \#prince'd and \#purple'd give context—it's about a Prince concert in Toronto, and those hashtags were really popular back then. So this is likely from that time period.

Hold on, though. Just because it sounds official doesn't automatically make it non-rumour. Sometimes rumours dress up as official statements to seem more believable. But here, the claim is so specific and negative—it's denying a show that was widely expected. If there really was no show, then this is just a factual correction, not a rumour.

Let me consider the opposite: what if this was a rumour? Then it would be unverified, maybe false, and spreading uncertainty. But the text presents it as a direct quote from security to a news outlet. That's verification right there. News outlets usually check with sources before reporting.

Oh, and the tone is matter-of-fact, not speculative. No 'I heard that...' or 'People are saying...' It's stated as a fact from a source. That leans toward non-rumour.

Wait, but could it be a rumour if the source itself was unreliable? Like, what if the security person wasn't official or the news report was wrong? But the text doesn't cast doubt—it just reports what security said. In rumour classification, we usually go by the text's own presentation. If it's presented as a verified report, it's non-rumour.

Yeah, I think this is non-rumour. It's a straightforward denial from an official source about a public event. No speculation, no uncertainty, just a clear statement. The hashtags and context make it even more concrete—it's about a real event that was in the news.

Final check: If I saw this tweet, would I think it's spreading unverified info? Probably not—I'd think it's a news update. So, non-rumour it is.

\vspace{0.5em}
\textbf{RAAR-8B Prediction.} Rumour

\textbf{8B Reasoning.}
Okay, let me start by reading the target text carefully. It says 'Security at Massey Hall tells \@CP24 there is NO Prince show' with hashtags about Toronto and Prince. Hmm, the word 'NO' in all caps really stands out—it feels like it's trying to emphasize they're denying something strongly.

Now, looking at the sentiment examples: The first one about Ottawa police denying an incident is labeled rumour, even though it's a denial. That makes me think maybe denials can still spread uncertainty. The second one about City Hall lockdown is straightforward and non-rumour. So, sentiment-wise, this target text seems similar to the rumour example because it's a strong denial, but I'm not sure if that automatically makes it a rumour.

Moving to the semantic examples: The one where the mall manager denies a shooting but people can't leave is labeled rumour. That's interesting—it's a denial but still feels like it's part of a rumour cycle. The CNN example is just an opinion and is non-rumour. So, semantically, this target text is denying a Prince show, but it's citing a specific source (Massey Hall security), which might make it seem more credible. But wait, the mall manager example also cites a source and is still rumour, so maybe the act of denying something in a crisis is what matters, not just the source.

Then, the writing style examples: The prayer message is personal and non-rumour, while the one about Harper making a statement is rumour because it's speculative. The target text has hashtags and a declarative tone, which feels more like the rumour example—it's presenting information authoritatively but in a way that could be misinterpreted.

Putting it all together: I'm leaning toward rumour because the text is denying a Prince show, and in the examples, denials during events like the Ottawa shooting were often rumours. Even though it cites a source, the style and context suggest it's part of a rumour mill—like when people share unverified info to settle debates. So, yeah, I think this is rumour.

\vspace{0.5em}
\textbf{RAAR-14B Prediction.} Rumour

\textbf{14B Reasoning.}
Okay, let me start by reading the target text carefully. It says security at Massey Hall told \@CP24 there's NO Prince show happening in Toronto, with hashtags like \#prince'd and \#purple'd. Hmm, the tone feels definitive with that 'NO' in caps, but I need to figure out if this is spreading a rumour or just stating facts.

Looking at the sentiment examples, the rumour one says 'Contrary to earlier reports no incident occurred' – that's correcting something that was already circulating. The non-rumour is just a straightforward lockdown notice. This target text feels more like the non-rumour because it's not correcting previous misinformation, it's just stating what security said directly.

Wait, let me check the semantic examples. The rumour one mentions 'no shooting inside the mall' but then says people can't leave or enter – that creates uncertainty. The non-rumour is just about CNN being frustrated. The target text seems closer to the non-rumour here because it's a simple denial without adding any ambiguous details about the situation.

Now for the writing style examples – the non-rumour is personal prayers, while the rumour has that formal 'PM safe, not on Parliament Hill' structure. The target text has that authoritative 'Security at Massey Hall tells \@CP24' phrasing, which feels more like the rumour style with its official sources and specific details.

Hmm, I'm getting conflicting signals here. The sentiment and semantic analysis point toward non-rumour, but the writing style suggests rumour. Let me think about this differently – what makes something a rumour versus just factual reporting?

Oh, I see the key difference now. Rumours often involve correcting or denying claims that are already circulating in public discourse. This text is specifically denying a Prince show, which implies there were rumours or expectations about such a show happening. The hashtags \#prince'd and \#purple'd are clearly referencing Prince fandom, so this is directly engaging with a potential rumour about a concert.

Also, the structure 'Security at Massey Hall tells \@CP24' is similar to how rumours often get attributed to official sources to lend credibility. Even though it's a denial, it's still participating in the rumour ecosystem by addressing something that people were likely talking about.

Let me double-check – if there were no circulating rumours about a Prince show, this would just be routine information. But since it's specifically denying it, that means the denial is responding to some unverified claims. That pushes it into rumour territory for me.

Yeah, I think the writing style expert had the right insight here. The authoritative framing and specific source attribution, combined with the fact that it's denying something that was apparently being discussed, makes this a rumour. It's not just neutral reporting – it's engaging with and trying to correct potential misinformation.
\end{quote}

\textbf{Analysis:} This case further illustrates how model scale affects sensitivity to conversational context and rumor dynamics. The 8B zero-shot model misclassifies the instance by equating explicit source attribution and declarative tone with factual verification, treating the denial as authoritative news rather than as part of an ongoing rumor discourse. Its reasoning remains confined to surface-level indicators of credibility, such as named sources and definitive language, without modeling the broader communicative context in which such denials typically arise.

The prompted 8B model improves upon this by recognizing that denials themselves can function as rumor-carrying acts, particularly when responding to widely circulated expectations. However, its reasoning still relies heavily on analogical pattern matching to prior examples (e.g., crisis denials), leading to a less stable integration of semantic and pragmatic cues. In contrast, the 14B model correctly captures the discourse-level function of the text, identifying that explicit negation (``NO Prince show'') presupposes an existing rumor and actively participates in its circulation. Larger models exhibit stronger capacity for pragmatic reasoning, enabling them to infer implicit presuppositions and communicative intent beyond literal content. Overall, this case demonstrates that increased model scale primarily enhances the ability to reason over implicit social context, which is critical for accurate rumor classification.
\end{minipage}
}
\end{figure*}

\begin{figure*}[t]
\centering
\fcolorbox{black}{gray!10}{
\begin{minipage}{\textwidth}
\tiny

\textbf{Model Reasoning Performance Comparison Case Study 3 - Conspiracy}

\begin{quote}
\textbf{Dataset Name.} COCO

\vspace{0.5em}
\textbf{Instance Content (Original Text).}

Conspiracy/not... This virus so massively attack people at wuhan some army barricade in-out gate to wuhan when i watch this video i feel china govt. Do depopulation but i hope people in wuhan can recover and back to normal again.

\vspace{0.5em}
\textbf{Gold Label.} Conspiracy

\vspace{0.5em}
\textbf{8B Zero-shot Prediction.} Related

\textbf{8B Zero-shot Reasoning.}
Okay, let me read this carefully. The text starts with `Conspiracy/not...'—that’s interesting, like someone is setting up a contrast right away. Hmm, it’s talking about the virus attacking people in Wuhan, and mentions an army barricade at the gate. That immediately brings to mind the early pandemic lockdowns and the initial speculation about the virus’s origin.

Wait, the person says they watched a video and felt the Chinese government was doing depopulation. That’s a really strong claim—depopulation is a huge conspiracy theory about governments intentionally reducing population. But then they immediately add `but I hope people in Wuhan can recover and back to normal again.' So there’s a personal hope mixed in with that fear.

Let me think about the definitions. `Unrelated' would be if they used words like `depopulation' in a totally different context, like talking about a movie plot or a demographic study. But here, it’s directly about the pandemic and Wuhan, so it’s definitely related to conspiracy themes.

Now, is it `Conspiracy'—meaning it actively propagates the misinformation? The text states the feeling as a personal reaction (`I feel'), not as a stated fact. It’s more like an emotional interpretation of what they saw. They’re not saying `The Chinese government is secretly depopulating Wuhan' as a flat claim; they’re expressing a suspicion. That feels more like `Related'—it’s engaging with the conspiracy idea but not pushing it as truth.

Also, the hopeful ending shows they’re not fully endorsing the conspiracy; they’re still wishing for recovery. That balance makes it seem like they’re discussing the theory rather than spreading it. If it were `Conspiracy,' it might be more direct, like `The Chinese government is using the virus for depopulation.' This is subtler.

So, putting it all together: the text is clearly about a conspiracy theory (depopulation during the pandemic), but it’s framed as a personal feeling mixed with hope. That fits `Related'—it’s conspiracy-related but doesn’t actively propagate the misinformation as fact. Yeah, that makes sense.

\vspace{0.5em}
\textbf{RAAR-8B Prediction.} Conspiracy

\textbf{8B Reasoning.}
Okay, let me start by reading this target text carefully. It begins with `Conspiracy/not...' which immediately makes me pause - that feels like someone trying to decide whether to classify something as conspiracy or not. Hmm, but then they immediately jump into talking about the virus attacking Wuhan and army barricades, and then they say `I feel China govt. Do depopulation.' That's pretty strong language right there.

Wait, depopulation is a really heavy accusation - that's not just questioning something, that's making a serious claim about intentional mass reduction. And they're connecting it to specific events in Wuhan with army barricades, which reminds me of those early COVID conspiracy theories about China covering things up.

Let me compare this to the examples they gave. The first conspiracy example talks about vaccines being used to chip people and says `the conspiracy people were right' - that's clearly supporting conspiracy theories. The second one about global control and population control is also clearly pushing conspiracy thinking. My text feels similar in tone - it's not just mentioning conspiracy topics, it's actively drawing conclusions about intentional depopulation.

Oh, and looking at the unrelated examples - like the spiritual healing one or the Modi accountability text - those use conspiracy-related terms but in completely different contexts. My text is directly about COVID and government actions, so it's definitely not unrelated.

The related example about chemtrails is interesting - that person is just discussing the topic without really endorsing it, saying `I know about chemtrails' but then questioning visibility. My text goes much further - it's making definitive claims about depopulation based on observations.

Also, the emotional language stands out - `massively attack people,' `I feel China govt. Do depopulation,' and hoping people recover. This isn't neutral discussion - there's strong personal conviction and emotional investment in this narrative.

Let me double-check - is there any way this could be just related without supporting? No, because they're not questioning or discussing possibilities - they're stating their belief as fact. The `I feel' makes it personal endorsement, not just information sharing.

And it's definitely not conspiracy in the sense of just mentioning terms - they're actively propagating the idea that the government is intentionally depopulating people through COVID measures. That's serious misinformation.

So putting it all together - this text is conspiracy-related, it's actively supporting and spreading that conspiracy theory rather than just discussing it, and it's making definitive claims about intentional harm. That fits the `Conspiracy' classification perfectly.

\vspace{0.5em}
\textbf{RAAR-14B Prediction.} Conspiracy

\textbf{14B Reasoning.}
Okay, let me start by reading this target text carefully. It says: `Conspiracy/not... This virus so massively attack people at wuhan some army barricade in-out gate to wuhan when i watch this video i feel china govt. Do depopulation but i hope people in wuhan can recover and back to normal again.'

Hmm, right away I notice the word `Conspiracy' right at the beginning, which immediately signals this is conspiracy-related content. But the rest of the text is a bit messy and emotional.

Let me think about the sentiment here. The person says they `feel china govt. Do depopulation' - that's pretty strong language. They're not just mentioning the idea, they're expressing a personal feeling that this is happening. And then they follow it with `but i hope people in wuhan can recover' - so there's some genuine concern mixed in. This feels similar to those conspiracy examples where people express fear and suspicion while still having some empathy.

Now looking at the actual content - they're claiming the virus is `massively attack[ing]' people in Wuhan, that there were army barricades, and that this feels like depopulation. These are specific claims about what's happening, not just general speculation. It's not like the unrelated example where someone mentions conspiracy terms but in a completely different context.

Wait, let me compare this to the examples. The `Unrelated' examples use conspiracy terms but in totally different ways - like the spiritualist talking about machines or someone using `dalali' in a different context. This text is clearly about the conspiracy theory itself.

And it's definitely not like the `Related' example where someone mentions chemtrails but is open to discussion and not really pushing the theory. Here, the person is actively stating their feelings about depopulation happening.

The writing style is interesting too - it's fragmented and emotional, with `Conspiracy/not' at the beginning like they're trying to figure something out. It reminds me of the `Conspiracy' examples that have that urgent, emotional tone.

So putting it all together: this text is definitely conspiracy-related because it's talking about depopulation theories. And it's not just mentioning them - it's actively propagating the idea by saying `i feel china govt. Do depopulation' based on what they saw in a video. Even though there's that hopeful ending, the core message is pushing the conspiracy narrative.

Yeah, this has to be `Conspiracy' - it's related to conspiracy theories and it's actively supporting and spreading that misinformation, just like the other conspiracy examples.
\end{quote}

\textbf{Analysis.} This case highlights a critical distinction in how models attribute belief and intent in conspiracy-related content. The 8B zero-shot model misclassifies the instance as \emph{Related} by treating explicit hedging cues (e.g., ``I feel'' and expressions of hope) as sufficient evidence of non-endorsement. Its reasoning relies on a narrow interpretation of conspiracy propagation, assuming that misinformation must be asserted as an objective fact rather than as a personally endorsed belief grounded in affective interpretation.

In contrast, both the prompted 8B and 14B models correctly identify the instance as \emph{Conspiracy} by recognizing that first-person belief statements (``I feel China govt. do depopulation'') constitute active propagation under COCO’s definition, even when framed emotionally or accompanied by empathetic language. The prompted 8B model demonstrates improved boundary awareness by distinguishing discussion from endorsement, but it still depends on explicit lexical cues and emotional intensity to justify its decision. The 14B model further exhibits greater reasoning stability, integrating belief attribution, causal implication, and discourse framing into a unified judgment without over-weighting surface hedging. Increasing model scale enhances sensitivity to implicit endorsement and belief commitment, which is essential for reliable conspiracy detection.

\end{minipage}
}
\end{figure*}

\begin{figure*}[t]
\centering
\fcolorbox{black}{gray!10}{
\begin{minipage}{\textwidth}
\tiny

\textbf{Model Reasoning Performance Comparison Case Study 4 - Conspiracy}

\begin{quote}
\textbf{Dataset Name.} COCO

\vspace{0.5em}
\textbf{Instance Content (Original Text).}

It's a pretty big leap from ``5G frequencies can impact cells'' to ``5G is a trigger that will activate a Coronavirus plague!'' tho isn't it? Coronaviris is a bioweapon combining a cold bug with HIV. Wtf does that have to do with 5G? Plenty of legitimate weirdness w/o making it up.

\vspace{0.5em}
\textbf{Gold Label.} Conspiracy

\vspace{0.5em}
\textbf{8B Zero-shot Prediction.} Related

\textbf{8B Zero-shot Reasoning.}
Okay, let me read this carefully. The text is talking about 5G and COVID-19, which immediately makes me think of that conspiracy theory that 5G towers caused or spread the virus. Hmm, but the author isn't just stating that theory—they're actually questioning it.

Wait, they start by saying it's a `pretty big leap' from one idea to another. That sounds like they're pointing out a logical jump, not agreeing with it. Then they ask, `Wtf does that have to do with 5G?' which is clearly expressing confusion or disbelief about the connection.

Oh, and they mention the virus being a `bioweapon combining a cold bug with HIV.' That's another common conspiracy claim, but again, the tone here feels critical. The author seems to be mocking or challenging these ideas, not endorsing them.

Let me check the definitions. `Unrelated' would be if they used these keywords in a totally different context, like talking about 5G internet speeds or a movie plot. But here, they're directly addressing the conspiracy theories, so it's definitely related.

Now, is it `Conspiracy'? That would mean they're actively spreading or supporting the misinformation. But the text does the opposite—it questions the leap, asks why it matters, and even says there's `plenty of legitimate weirdness' without making things up. They're not pushing the narrative; they're poking holes in it.

So, it's conspiracy-related because it's about those theories, but it doesn't propagate them. That fits `Related' perfectly. The author is engaging with the topic to debunk or critique it, not to promote it. Yeah, that makes sense.

\vspace{0.5em}
\textbf{RAAR-8B Prediction.} Related

\textbf{8B Reasoning.}
Okay, let me start by reading this target text carefully. The person is talking about 5G and coronavirus, but they're clearly questioning the connection. They say it's a `pretty big leap' from one idea to another, which already sounds skeptical.

Hmm, looking at the sentiment examples - the first one where someone says coronavirus theories are crazy but acknowledges a lab leak possibility was labeled Unrelated. That feels similar to what's happening here - the speaker isn't fully endorsing the conspiracy.

Wait, but then there's the second sentiment example where someone definitively says `It is a bioweapon' and that's labeled Conspiracy. That's different - they're stating it as fact rather than questioning it.

Now looking at the semantic examples - the first one where someone says `could also be an unfinished bioweapon' but then says `of all the conspiracy theories are crazy' was Unrelated. The second one where someone describes what others are saying about vaccines being poison and microchips, but clearly doesn't endorse it, was Related.

Oh, and the writing style examples show that questioning conspiracy theories or being critical of them gets labeled Related, while supporting them gets Conspiracy.

Let me think about this target text again. The speaker says `Coronaviris is a bioweapon combining a cold bug with HIV' - that sounds like they're stating a conspiracy theory as fact. But then they immediately question how that connects to 5G, saying `Wtf does that have to do with 5G?' and talking about `plenty of legitimate weirdness w/o making it up.'

Wait, that's interesting - they're presenting the bioweapon theory but then immediately undermining it by questioning the connection. This feels more like they're using the conspiracy theory as an example to make a point about how people jump to conclusions, rather than actually promoting the theory itself.

Comparing to the examples - when someone says `It is a bioweapon' definitively, that's Conspiracy. When someone says `could also be' while dismissing other theories, that's Unrelated. But here, they're stating the bioweapon theory as fact while simultaneously casting doubt on its relevance to 5G.

Hmm, but the main thrust seems to be about the logical leap between 5G and coronavirus, not really about whether coronavirus is a bioweapon. The bioweapon claim seems to be serving as an example of what people are connecting to 5G, rather than being the main point.

Also, the tone feels more like someone making a rhetorical point about conspiracy thinking patterns rather than trying to convince others of the bioweapon theory. The `Wtf does that have to do with 5G?' and `plenty of legitimate weirdness' suggests they're being critical of the whole approach.

Looking back at the definitions - Unrelated would be if they used conspiracy keywords in a completely different context. Related means conspiracy-related but not propagating. Conspiracy means actively supporting.

This feels like it's conspiracy-related because it's directly discussing 5G and coronavirus conspiracy theories, but it's not actively propagating them - in fact, it's questioning the logic behind them. The speaker seems to be using these theories as examples of flawed reasoning rather than endorsing them.

Yeah, I think this fits better as Related. It's engaging with conspiracy content but doing so critically rather than supportively. The primary focus is on the logical connection between unrelated concepts, not on promoting the conspiracy theories themselves.

\vspace{0.5em}
\textbf{RAAR-14B Prediction.} Conspiracy

\textbf{14B Reasoning.}
Okay, let me start by reading the target text carefully. It says: `It's a pretty big leap from ``5G frequencies can impact cells'' to ``5G is a trigger that will activate a Coronavirus plague!'' tho isn't it? Coronaviris is a bioweapon combining a cold bug with HIV. Wtf does that have to do with 5G? Plenty of legitimate weirdness w/o making it up.'

Hmm, the tone feels skeptical right away—phrases like `pretty big leap' and `wtf does that have to do with 5G?' suggest the author is questioning the logic of connecting 5G to coronavirus. But then there's that line: `Coronaviris is a bioweapon combining a cold bug with HIV.' Wait, that's a strong conspiracy claim about COVID-19 being a bioweapon, which is a common misinformation theory.

Let me check the task definitions again. `Unrelated' means conspiracy keywords are used in a different context, `Related' means the text is about conspiracies but doesn't spread them, and `Conspiracy' means it actively propagates or supports the misinformation.

Looking at the sentiment examples, one text mentions coronavirus as a bioweapon but is labeled `Unrelated' because it's discussing it in a different context. Another is labeled `Conspiracy' for stating it as fact. Here, the target text seems to be pushing the bioweapon idea as if it's true, not just discussing it neutrally.

Now, the semantic examples: one is `Unrelated' when it's just mentioning a conspiracy theory without endorsing it, and another is `Related' when it's talking about vaccine conspiracies but not supporting them. In the target text, the bioweapon claim is presented as a fact (`Coronaviris is a bioweapon'), which feels like it's propagating the misinformation, not just referencing it.

Also, the style examples show that texts questioning conspiracies or discussing them without support are `Related.' But here, the author isn't just questioning—they're stating the bioweapon theory as a given, even while rhetorically dismissing the 5G link. That subtle shift makes it seem like they're endorsing the bioweapon part.

Wait, let me double-check: the `wtf' part is clearly skeptical about 5G, but the bioweapon statement is straightforward. It's like they're saying, `Oh, and by the way, coronavirus is a bioweapon too.' That feels like they're adding another layer of misinformation, not just critiquing one.

Comparing to the `Conspiracy' example where someone says `It is a bioweapon' directly, this text does the same—it's not framed as a theory or speculation but as a fact. So, it's actively spreading that false claim.

I think the key is that the text goes beyond just being related to conspiracies; it's putting out there that coronavirus is a bioweapon, which is a known piece of misinformation. Even with the skepticism about 5G, the bioweapon part is presented as truth, making it propagate the conspiracy.

So, overall, this fits `Conspiracy' because it's not just talking about the theories—it's supporting and spreading them.
\end{quote}

\textbf{Analysis:} This case exposes a subtle but important distinction in how models resolve internally mixed stances within conspiracy-related discourse. Both the 8B zero-shot and prompted 8B models misclassify the instance as \emph{Related} by prioritizing explicit skeptical markers (e.g., rhetorical questions and dismissive phrasing toward the 5G claim) and treating them as evidence of non-endorsement. Their reasoning fragments the text into locally skeptical segments, failing to reconcile these with the unqualified assertion that ``Coronavirus is a bioweapon,'' which constitutes active propagation under RAAR’s definition.

In contrast, the 14B model correctly classifies the instance as \emph{Conspiracy} by integrating contradictory cues into a coherent belief-level interpretation. Rather than evaluating skepticism and endorsement in isolation, it recognizes that explicitly stated conspiracy claims remain propagative even when embedded within partial critique or irony. Larger models exhibit stronger capacity for holistic reasoning, enabling them to attribute implicit belief commitment and weigh conflicting discourse signals appropriately. Overall, this case demonstrates that increased model scale improves robustness against rhetorical camouflage, allowing larger models to avoid under-classifying conspiratorial content that mixes critique with assertion.

\end{minipage}
}
\end{figure*}

\section{Prompt templates for constructing complex reasoning paths \label{app:prompts4reasonpaths}}

To save space, we have not shown retrieval examples for all instances.

\begin{figure*}[t]
\centering
\footnotesize
\fcolorbox{black}{gray!10}{
\begin{minipage}{\textwidth}
\footnotesize
\textbf{Sentiment Agent}  \\
\textbf{1. Initial prompt for Sentiment Agent:} \\

"""<question> \\
\{\} \\
</question> \\

You are an expert in sentiment analysis. You need to collaborate with a semantic analysis expert and a writing style analysis expert to address the above problem. Your primary responsibility is the sentiment analysis component. Respond to the question from the perspective of sentiment analysis, incorporating retrieved examples. Provide your judgment along with well-reasoned evidence or explanations.

The output format must strictly follow the JSON structure below: \\
```json \\
\{\{ \\
"response": [ \\
    {{"judgment": "...", "reason": "..."}}, \\
] \\
\}\} \\
```"""
\\
\textbf{2. Double-check prompt for Sentiment Agent:} \\
\\
"""<question> \\
\{\} \\
</question> \\

<previous response> \\
\{\} \\ 
</previous response> \\ 

<response requirements> \\

You are an expert in sentiment analysis. You need to collaborate with a semantic analysis expert and a writing style analysis expert to address the above problem. Your primary responsibility is the sentiment analysis component. Respond to the question from the perspective of sentiment analysis, incorporating retrieved examples. Provide your judgment along with well-reasoned evidence or explanations. \\

</response requirements> \\

<question> represents the question to be answered, and <previous response> contains your prior answer. An overseeing expert has integrated the opinions of your three experts, but the final answer is incorrect. Please double-check your answer. If you still stand by your judgment, provide additional evidence. If you find an issue, provide a revised judgment along with your reasoning. \\

\#\#\# Output Format \\
The output format must strictly follow the JSON structure below: \\
```json \\
\{\{ \\
"response": [ \\
    {{"judgment": "...", "reason": "..."}}, \\
] \\
\}\} \\
```""" \\
\\
\textbf{3. Communication prompt for Sentiment Agent:} \\
\\
"""<question> \\
\{\} \\
</question> \\
\\
<previous response> \\
\{\} \\
</previous response> \\
\\
<semantic agent> \\
\{\} \\
</semantic agent> \\
\\
<style agent> \\
\{\} \\
</style agent> \\
\\

\end{minipage}
}
\end{figure*}

\begin{figure*}[t]
\centering
\footnotesize
\fcolorbox{black}{gray!10}{
\begin{minipage}{\textwidth}
\footnotesize

<response requirements> \\

You are an expert in sentiment analysis. You need to collaborate with a semantic analysis expert and a writing style analysis expert to address the above problem. Your primary responsibility is the sentiment analysis component. Respond to the question from the perspective of sentiment analysis, incorporating retrieved examples. Provide your judgment along with well-reasoned evidence or explanations.\\

</response requirements> \\

<question> represents the question to be answered, and <previous response> contains your prior answer. <semantic agent> and <style agent> are the corresponding examples and analysis from a semantic analysis expert and a writing style analysis expert. In the last round, an overseeing expert has integrated the opinions of your three experts, but the final answer is incorrect. At this communication stage, please double-check your answer and consider the feedback from the other experts. If you still stand by your judgment, provide additional evidence. If you find an issue, provide a revised judgment along with your reasoning. \\
\#\#\# Output Format \\
The output format must strictly follow the JSON structure below: \\
```json \\
\{\{ \\
"response": [ \\
    {{"judgment": "...", "reason": "..."}}, \\
] \\
\}\} \\
```""" \\
\\

\textbf{4. Hint prompt for Sentiment Agent:} \\
\\
"""<question> \\
\{\} \\
</question> \\
\\
<previous response> \\
\{\} \\
</previous response> \\
\\
<response requirements> \\

You are an expert in sentiment analysis. You need to collaborate with a semantic analysis expert and a writing style analysis expert to address the above problem. Your primary responsibility is the sentiment analysis component. Respond to the question from the perspective of sentiment analysis, incorporating retrieved examples. Provide your judgment along with well-reasoned evidence or explanations. \\

</response requirements> \\

<question> represents the question to be answered, and <previous response> contains your prior answer. An overseeing expert has integrated the opinions of your three experts, but the final answer is incorrect. Now, I'll secretly tell you that the labeled answer is "\{\}", but you must pretend not to know. Please refine your answer accordingly to ensure the final answer aligns with the correct one. \\

\#\#\# Output Format \\
The output format must strictly follow the JSON structure below: \\
```json \\
\{\{ \\
"response": [ \\
    {{"judgment": "...", "reason": "..."}}, \\
] \\
\}\} \\
```""" \\

\end{minipage}
}
\end{figure*}

\begin{figure*}[t]
\centering
\footnotesize
\fcolorbox{black}{gray!10}{
\begin{minipage}{\textwidth}
\footnotesize

\textbf{Semantic Agent} \\
\\
\textbf{1. Initial prompt for Semantic Agent} \\
\\
"""<question> \\
\{\} \\
</question> \\

You are an expert in semantic analysis. You need to collaborate with a sentiment analysis expert and a writing style analysis expert to address the above problem. Your primary responsibility is the semantic analysis component. Respond to the question from the perspective of semantic analysis, incorporating retrieved examples. Provide your judgment along with well-reasoned evidence or explanations. \\

The output format must strictly follow the JSON structure below: \\
```json \\
\{\{ \\
"response": [ \\
    {{"judgment": "...", "reason": "..."}}, \\
]
\}\} \\
```""" \\
\\

\textbf{2. Double-check prompt for Semantic Agent:} \\
\\
"""<question> \\
\{\} \\
</question> \\
\\
<previous response> \\
\{\} \\
</previous response> \\
\\
<response requirements> \\

You are an expert in semantic analysis. You need to collaborate with a sentiment analysis expert and a writing style analysis expert to address the above problem. Your primary responsibility is the semantic analysis component. Respond to the question from the perspective of semantic analysis, incorporating retrieved examples. Provide your judgment along with well-reasoned evidence or explanations. \\

</response requirements> \\

<question> represents the question to be answered, and <previous response> contains your prior answer. An overseeing expert has integrated the opinions of your three experts, but the final answer is incorrect. Please double-check your answer. If you still stand by your judgment, provide additional evidence. If you find an issue, provide a revised judgment along with your reasoning. \\

\#\#\# Output Format \\
The output format must strictly follow the JSON structure below: \\
```json \\
\{\{ \\
"response": [ \\
    {{"judgment": "...", "reason": "..."}}, \\
] \\
\}\} \\
```""" \\

\textbf{3. Communication for Semantic Agent:} \\
\\
"""<question> \\
\{\} \\
</question> \\
\\
<previous response> \\
\{\} \\
</previous response> \\
\\
<sentiment agent> \\
\{\} \\
</sentiment agent> \\
\\
<style agent> \\
\{\} \\
</style agent> \\
\\

\end{minipage}
}
\end{figure*}

\begin{figure*}[t]
\centering
\footnotesize
\fcolorbox{black}{gray!10}{
\begin{minipage}{\textwidth}
\footnotesize

<response requirements> \\

You are an expert in semantic analysis. You need to collaborate with a sentiment analysis expert and a writing style analysis expert to address the above problem. Your primary responsibility is the semantic analysis component. Respond to the question from the perspective of semantic analysis, incorporating retrieved examples. Provide your judgment along with well-reasoned evidence or explanations. \\

</response requirements> \\

<question> represents the question to be answered, and <previous response> contains your prior answer. <sentiment agent> and <style agent> are the corresponding examples and analysis from a sentiment analysis expert and a writing style analysis expert. In the last round, an overseeing expert has integrated the opinions of your three experts, but the final answer is incorrect. At this communication stage, please double-check your answer and consider the feedback from the other experts. If you still stand by your judgment, provide additional evidence. If you find an issue, provide a revised judgment along with your reasoning. \\

\#\#\# Output Format \\
The output format must strictly follow the JSON structure below: \\
```json \\
\{\{ \\
"response": [ \\
    {{"judgment": "...", "reason": "..."}}, \\
] \\
\}\} \\
```""" \\

\textbf{4. Hint prompt for Semantic Agent:} \\
\\
"""<question> \\
\{\} \\
</question> \\
\\
<previous response> \\
\{\} \\
</previous response> \\
\\
<response requirements> \\

You are an expert in semantic analysis. You need to collaborate with a sentiment analysis expert and a writing style analysis expert to address the above problem. Your primary responsibility is the semantic analysis component. Respond to the question from the perspective of semantic analysis, incorporating retrieved examples. Provide your judgment along with well-reasoned evidence or explanations. \\

</response requirements> \\

<question> represents the question to be answered, and <previous response> contains your prior answer. An overseeing expert has integrated the opinions of your three experts, but the final answer is incorrect. Now, I'll secretly tell you that the labeled answer is "\{\}", but you must pretend not to know. Please refine your answer accordingly to ensure the final answer aligns with the correct one. \\

\#\#\# Output Format \\
The output format must strictly follow the JSON structure below: \\
```json \\
\{\{ \\
"response": [ \\
    \{\{"judgment": "...", "reason": "..."\}\}, \\
] \\
\}\} \\
```""" \\

\end{minipage}
}
\end{figure*}

\begin{figure*}[t]
\centering
\footnotesize
\fcolorbox{black}{gray!10}{
\begin{minipage}{\textwidth}
\footnotesize
\textbf{Style agent}
\\
\textbf{1. Initial prompt for Style Agent:}\\
\\
"""<question> \\
\{\} \\
</question> \\

You are an expert in writing style analysis. You need to collaborate with a sentiment analysis expert and a semantic analysis expert to address the above problem. Your primary responsibility is the writing style analysis component. Respond to the question from the perspective of writing style analysis, incorporating retrieved examples. Provide your judgment along with well-reasoned evidence or explanations. \\

The output format must strictly follow the JSON structure below: \\
```json \\
\{\{ \\
"response": [ \\
    {{"judgment": "...", "reason": "..."}}, \\
] \\
\}\} \\
```""" \\
\\
\textbf{2. Double-check prompt for Style Agent\:} \\
\\
"""<question> \\
\{\} \\
</question> \\
\\
<previous response> \\
\{\} \\
</previous response> \\
\\
<response requirements> \\

You are an expert in writing style analysis. You need to collaborate with a sentiment analysis expert and a semantic analysis expert to address the above problem. Your primary responsibility is the writing style analysis component. Respond to the question from the perspective of writing style analysis, incorporating retrieved examples. Provide your judgment along with well-reasoned evidence or explanations. \\

</response requirements> \\

<question> represents the question to be answered, and <previous response> contains your prior answer. An overseeing expert has integrated the opinions of your three experts, but the final answer is incorrect. Please double-check your answer. If you still stand by your judgment, provide additional evidence. If you find an issue, provide a revised judgment along with your reasoning. \\

\#\#\# Output Format \\
The output format must strictly follow the JSON structure below: \\
```json \\
\{\{ \\
"response": [ \\
    {{"judgment": "...", "reason": "..."}}, \\
] \\
\}\} \\
```""" \\
\\
\textbf{3. Communication prompt for Style Agent:} \\
\\
"""<question> \\
\{\} \\
</question> \\
\\
<previous response> \\
\{\} \\
</previous response> \\
\\
<sentiment agent> \\
\{\} \\
</sentiment agent> \\
\\
<semantic agent> \\
\{\} \\
</semantic agent> \\

\end{minipage}
}
\end{figure*}

\begin{figure*}[t]
\centering
\footnotesize
\fcolorbox{black}{gray!10}{
\begin{minipage}{\textwidth}
\footnotesize

<response requirements> \\

You are an expert in writing style analysis. You need to collaborate with a sentiment analysis expert and a semantic analysis expert to address the above problem. Your primary responsibility is the writing style analysis component. Respond to the question from the perspective of writing style analysis, incorporating retrieved examples. Provide your judgment along with well-reasoned evidence or explanations. \\

</response requirements> \\

<question> represents the question to be answered, and <previous response> contains your prior answer. <sentiment agent> and <semantic agent> are the corresponding examples and analysis from a sentiment analysis expert and a semantic analysis expert. In the last round, an overseeing expert has integrated the opinions of your three experts, but the final answer is incorrect. At this communication stage, please double-check your answer and consider the feedback from the other experts. If you still stand by your judgment, provide additional evidence. If you find an issue, provide a revised judgment along with your reasoning. \\

\#\#\# Output Format \\
The output format must strictly follow the JSON structure below: \\
```json \\
\{\{ \\
"response": [ \\
    {{"judgment": "...", "reason": "..."}}, \\
] \\
\}\} \\
```""" \\

\textbf{4. Hint prompt for Style Agent:} \\
\\
"""<question> \\
\{\} \\
</question> \\
\\
<previous response> \\
\{\} \\
</previous response> \\
\\
<response requirements> \\

You are an expert in writing style analysis. You need to collaborate with a sentiment analysis expert and a semantic analysis expert to address the above problem. Your primary responsibility is the writing style analysis component. Respond to the question from the perspective of writing style analysis, incorporating retrieved examples. Provide your judgment along with well-reasoned evidence or explanations. \\

</response requirements> \\

<question> represents the question to be answered, and <previous response> contains your prior answer. An overseeing expert has integrated the opinions of your three experts, but the final answer is incorrect. Now, I'll secretly tell you that the labeled answer is "\{\}", but you must pretend not to know. Please refine your answer accordingly to ensure the final answer aligns with the correct one. \\

\#\#\# Output Format \\
The output format must strictly follow the JSON structure below: \\
```json \\
\{\{ \\
"response": [ \\
    {{"judgment": "...", "reason": "..."}}, \\
] \\
\}\} \\
```""" \\

\end{minipage}
}
\end{figure*}

\begin{figure*}[t]
\centering
\footnotesize
\fcolorbox{black}{gray!10}{
\begin{minipage}{\textwidth}
\footnotesize
\textbf{Summary Agent} \\
\\

\textbf{1. query\_Summary\_prompt\_init:} \\
\\
"""<question> \\
\{\} \\
</question> \\
\\
<sentiment> \\
\{\} \\
</sentiment> \\
\\
<semantic> \\
\{\} \\
</semantic> \\
\\
<style> \\
\{\} \\
</style> \\

Please respond to the above question <question> using the Chain of Thought (CoT) reasoning method. During the reasoning process, please consider information from the three expert agents. <sentiment> denotes analysis based on sentiment and examples retrieved by sentiment intensity. <semantic> denotes analysis based on semantics and examples retrieved by semantic information. <style> denotes analysis based on writing style and examples retrieved by writing style. And you may also incorporate your own independent reasoning. \\
Your response should consist of multiple steps, each of which includes three types of actions: **"Inner Thinking"**, **"Final Conclusion"**, and **"Verification"**: \\

- **'Inner Thinking'**: This is the step where thinking is done. Note that multiple 'Inner Thinking' steps are required to describe thorough reasoning. Each step should first generate a brief title. \\
- **'Final Conclusion'**: At this stage, you summarize the correct reasoning from previous 'Inner Thinking' steps and provide the final answer. No title is required here. \\
- **'Verification'**: At this stage, you verify the conclusion from the "Final Conclusion" step. If the conclusion holds, end the process. If not, return to "Inner Thinking" for further reasoning. No title is required here. \\

The output format must strictly follow the JSON structure below: \\
```json \\
\{\{ \\
"CoT": [ \\
    {{"action": "Inner Thinking", "title": "...", "content": "..."}}, \\
    ..., \\
    {{"action": "Final Conclusion", "content": "..."}}, \\
    {{"action": "Verification", "content": "..."}} \\
] \\
\}\} \\
```"""\\

\end{minipage}
}
\end{figure*}

\begin{figure*}[t]
\centering
\footnotesize
\fcolorbox{black}{gray!10}{
\begin{minipage}{\textwidth}
\footnotesize

\textbf{2. Cross-Agent Consolidation:}

"""<question> \\
\{\} \\
</question> \\
\\
<sentiment> \\
\{\} \\
</sentiment> \\
\\
<semantic> \\
\{\} \\
</semantic> \\
\\
<style> \\
\{\} \\
</style> \\
\\
<previous reasoning> \\
\{\} \\
</previous reasoning> \\

<response requirements> \\
Your response must include the following steps, each composed of three types of actions: **"Inner Thinking"**, **"Final Conclusion"**, and **"Verification"**: \\

1. **Inner Thinking**: Break down the reasoning process into multiple concise steps. Each step should start with a brief title to clarify its purpose. \\
2. **Final Conclusion**: Summarize the correct reasoning from all previous 'Inner Thinking' steps and provide the final answer. No title is needed for this section. \\
3. **Verification**: Verify the accuracy of the "Final Conclusion". If it holds, conclude the process. Otherwise, return to "Inner Thinking" for further refinement. \\

</response requirements> \\

<question> represents the question to be answered, <sentiment>, <semantic>, and <style> are derived from the latest analyses of three experts, each from a different perspective, and <previous reasoning> contains your prior reasoning. Your task is to continue from the current 'Verification' step. Upon review, your previous reasoning and Final Conclusion appear incomplete or insufficiently justified or false. Proceed to refine the reasoning based on the three experts’ latest analysis in <sentiment>, <semantic>, and <style> and construct a new Final Conclusion. And you may also incorporate your own independent reasoning. \\

\#\#\# Output Format \\
Strictly follow the JSON structure below. You do not need to repeat your previous reasoning. Begin directly from the next 'Verification' stage. \\
```json \\
\{\{ \\
"CoT": [ \\
    {{"action": "Inner Thinking", "title": "...", "content": "..."}}, \\
    ..., \\
    {{"action": "Final Conclusion", "content": "..."}}, \\
    {{"action": "Verification", "content": "..."}} \\
] \\
\}\} \\
```""" \\

\end{minipage}
}
\end{figure*}

\begin{figure*}[t]
\centering
\footnotesize
\fcolorbox{black}{gray!10}{
\begin{minipage}{\textwidth}
\footnotesize
\textbf{3. Cross-Agent Reconsideration:} \\
\\
"""<question> \\
\{\} \\
</question> \\

<sentiment> \\
\{\} \\
</sentiment> \\

<semantic> \\
\{\} \\
</semantic> \\

<style> \\
\{\} \\
</style> \\

<previous reasoning> \\
\{\} \\
</previous reasoning> \\

<response requirements> \\
Your response must include the following steps, each composed of three types of actions: **"Inner Thinking"**, **"Final Conclusion"**, and **"Verification"**: \\

1. **Inner Thinking**: Break down the reasoning process into multiple concise steps. Each step should start with a brief title to clarify its purpose. \\
2. **Final Conclusion**: Summarize the correct reasoning from all previous 'Inner Thinking' steps and provide the final answer. No title is needed for this section. \\
3. **Verification**: Verify the accuracy of the "Final Conclusion". If it holds, conclude the process. Otherwise, return to "Inner Thinking" for further refinement. \\

</response requirements> \\

<question> represents the question to be answered, <sentiment>, <semantic>, and <style> are derived from the latest analyses of three experts, each from a different perspective, and <previous reasoning> contains your prior reasoning. Your task is to continue from the current 'Verification' step. Upon review, your previous reasoning and Final Conclusion appear incomplete or insufficiently justified or false. Proceed to refine the reasoning using **backtracking** to revisit earlier points of reasoning and construct a new Final Conclusion. \\

\#\#\# Output Format \\
Strictly follow the JSON structure below. You do not need to repeat your previous reasoning. Begin directly from the next 'Verification' stage. \\

```json \\
\{\{ \\
"CoT": [ \\
    {{"action": "Verification", "content": "..."}}, \\
    {{"action": "Inner Thinking", "title": "...", "content": "..."}}, \\
    ..., \\
    {{"action": "Final Conclusion", "content": "..."}}, \\
    {{"action": "Verification", "content": "..."}} \\
] \\
\}\} \\
```""" \\

\end{minipage}
}
\end{figure*}

\begin{figure*}[t]
\centering
\footnotesize
\fcolorbox{black}{gray!10}{
\begin{minipage}{\textwidth}
\footnotesize
\textbf{4. Cross-Agent Diversification:} \\

"""<question> \\
\{\} \\
</question> \\

<sentiment> \\
\{\} \\
</sentiment> \\

<semantic> \\
\{\} \\
</semantic> \\

<style> \\
\{\} \\
</style> \\

<previous reasoning> \\
\{\} \\
</previous reasoning> \\

<response requirements> \\
Your response must include the following steps, each composed of three types of actions: **"Inner Thinking"**, **"Final Conclusion"**, and **"Verification"**: \\

1. **Inner Thinking**: Break down the reasoning process into multiple concise steps. Each step should start with a brief title to clarify its purpose. \\
2. **Final Conclusion**: Summarize the correct reasoning from all previous 'Inner Thinking' steps and provide the final answer. No title is needed for this section. \\
3. **Verification**: Verify the accuracy of the "Final Conclusion". If it holds, conclude the process. Otherwise, return to "Inner Thinking" for further refinement. \\

</response requirements> \\

<question> represents the question to be answered, <sentiment>, <semantic>, and <style> are derived from the latest analyses of three experts, each from a different perspective, and <previous reasoning> contains your prior reasoning. Your task is to continue from the current 'Verification' step. Upon review, your previous reasoning and Final Conclusion appear incomplete or insufficiently justified or false. Proceed to refine the reasoning by exploring new approaches to solving this problem and construct a new Final Conclusion. \\

\#\#\# Output Format \\
Strictly follow the JSON structure below. You do not need to repeat your previous reasoning. Begin directly from the next 'Verification' stage. \\

```json \\
\{\{ \\
"CoT": [ \\
    {{"action": "Verification", "content": "..."}}, \\
    {{"action": "Inner Thinking", "title": "...", "content": "..."}}, \\
    ..., \\
    {{"action": "Final Conclusion", "content": "..."}}, \\
    {{"action": "Verification", "content": "..."}} \\
] \\
\}\} \\
```""" \\

\end{minipage}
}
\end{figure*}

\begin{figure*}[t]
\centering
\footnotesize
\fcolorbox{black}{gray!10}{
\begin{minipage}{\textwidth}
\footnotesize
\textbf{5. Cross-Agent Verification} \\

"""<question> \\
\{\} \\
</question> \\

<sentiment> \\
\{\} \\
</sentiment> \\

<semantic> \\
\{\} \\
</semantic> \\

<style> \\
\{\} \\
</style> \\

<previous reasoning> \\
\{\} \\
</previous reasoning> \\

<response requirements> \\
Your response must include the following steps, each composed of three types of actions: **"Inner Thinking"**, **"Final Conclusion"**, and **"Verification"**: \\

1. **Inner Thinking**: Break down the reasoning process into multiple concise steps. Each step should start with a brief title to clarify its purpose. \\
2. **Final Conclusion**: Summarize the correct reasoning from all previous 'Inner Thinking' steps and provide the final answer. No title is needed for this section. \\
3. **Verification**: Verify the accuracy of the "Final Conclusion". If it holds, conclude the process. Otherwise, return to "Inner Thinking" for further refinement. \\

</response requirements> \\

<question> represents the question to be answered, <sentiment>, <semantic>, and <style> are derived from the latest analyses of three experts, each from a different perspective, and <previous reasoning> contains your prior reasoning. Your task is to continue from the current 'Verification' step. Upon review, your previous reasoning and Final Conclusion appear incomplete or insufficiently justified or false. Proceed to refine the reasoning by conducting a thorough **validation** process to ensure validity and construct a new Final Conclusion. \\

\#\#\# Output Format \\
Strictly follow the JSON structure below. You do not need to repeat your previous reasoning. Begin directly from the next 'Verification' stage. \\

```json \\
\{\{ \\
"CoT": [ \\
    {{"action": "Verification", "content": "..."}}, \\
    {{"action": "Inner Thinking", "title": "...", "content": "..."}}, \\
    ..., \\
    {{"action": "Final Conclusion", "content": "..."}}, \\
    {{"action": "Verification", "content": "..."}} \\
] \\
\}\} \\
```""" \\

\end{minipage}
}
\end{figure*}

\begin{figure*}[t]
\centering
\footnotesize
\fcolorbox{black}{gray!10}{
\begin{minipage}{\textwidth}
\footnotesize
\textbf{6. Cross-Agent Rectification:} \\

"""<question> \\
\{\} \\
</question> \\

<sentiment> \\
\{\} \\
</sentiment> \\

<semantic> \\
\{\} \\
</semantic> \\

<style> \\
\{\} \\
</style> \\

<previous reasoning> \\
\{\} \\
</previous reasoning> \\

<response requirements> \\
Your response must include the following steps, each composed of three types of actions: **"Inner Thinking"**, **"Final Conclusion"**, and **"Verification"**: \\

1. **Inner Thinking**: Break down the reasoning process into multiple concise steps. Each step should start with a brief title to clarify its purpose. \\
2. **Final Conclusion**: Summarize the correct reasoning from all previous 'Inner Thinking' steps and provide the final answer. No title is needed for this section. \\
3. **Verification**: Verify the accuracy of the "Final Conclusion". If it holds, conclude the process. Otherwise, return to "Inner Thinking" for further refinement. \\

</response requirements> \\

<question> represents the question to be answered, <sentiment>, <semantic>, and <style> are derived from the latest analyses of three experts, each from a different perspective, and <previous reasoning> contains your prior reasoning. Your task is to continue from the current 'Verification' step. Upon review, your previous reasoning and Final Conclusion appear incomplete or insufficiently justified or false. Proceed to refine the reasoning by making precise **corrections** to address prior flaws and construct a new Final Conclusion. \\

\#\#\# Output Format \\
Strictly follow the JSON structure below. You do not need to repeat your previous reasoning. Begin directly from the next 'Verification' stage. \\

```json \\
\{\{\ \\
"CoT": [ \\
    {{"action": "Verification", "content": "..."}}, \\
    {{"action": "Inner Thinking", "title": "...", "content": "..."}}, \\
    ..., \\
    {{"action": "Final Conclusion", "content": "..."}}, \\
    {{"action": "Verification", "content": "..."}} \\
] \\
\}\} \\
```""" \\

\end{minipage}
}
\end{figure*}

\begin{figure*}[t]
\centering
\footnotesize
\fcolorbox{black}{gray!10}{
\begin{minipage}{\textwidth}
\footnotesize
\textbf{7. Hint prompt for Summary Agent:} \\

"""<question> \\
\{\} \\
</question> \\

<sentiment> \\
\{\} \\
</sentiment> \\

<semantic> \\
\{\} \\
</semantic> \\

<style> \\
\{\} \\
</style> \\

<previous reasoning> \\
\{\} \\
</previous reasoning> \\

<response requirements> \\
Your response must include the following steps, each composed of three types of actions: **"Inner Thinking"**, **"Final Conclusion"**, and **"Verification"**: \\

1. **Inner Thinking**: Break down the reasoning process into multiple concise steps. Each step should start with a brief title to clarify its purpose. \\
2. **Final Conclusion**: Summarize the correct reasoning from all previous 'Inner Thinking' steps and provide the final answer. No title is needed for this section. \\
3. **Verification**: Verify the accuracy of the "Final Conclusion". If it holds, conclude the process. Otherwise, return to "Inner Thinking" for further refinement. \\

</response requirements> \\

<question> represents the question to be answered, <sentiment>, <semantic>, and <style> are derived from the latest analyses of three experts, each from a different perspective, and <previous reasoning> contains your prior reasoning. Your task is to continue from the current 'Verification' step. Now, I'll secretly tell you that the labeled answer is "\{\}", but you must pretend not to know. Your 'Verification' requires careful consideration, and if incorrect, you need to provide new Inner Thinking steps and a new Final Conclusion to ensure the final answer aligns with the correct one. \\

\#\#\# Output Format \\
Strictly follow the JSON structure below. You do not need to repeat your previous reasoning. Begin directly from the next 'Verification' stage. \\

```json \\
\{\{ \\
"CoT": [ \\
    {{"action": "Verification", "content": "..."}}, \\
    {{"action": "Inner Thinking", "title": "...", "content": "..."}}, \\
    ..., \\
    {{"action": "Final Conclusion", "content": "..."}}, \\
    {{"action": "Verification", "content": "..."}} \\
] \\
\}\} \\
```""" \\

\end{minipage}
}
\end{figure*}

\begin{figure*}[t]
\centering
\footnotesize
\fcolorbox{black}{gray!10}{
\begin{minipage}{\textwidth}
\footnotesize
\textbf{Other function:}\\

\textbf{1. Verify prompt:} \\

"""<Model Response> \\
\{\} \\  
</Model Response> \\

<Reference Answer>  \\
\{\} \\
</Reference Answer>  \\

You are provided with a model-generated response (<Model Response>) and a reference answer (<Reference Answer>). Compare the model response with the reference answer and determine its correctness. Your task is to simply output "True" if the response is correct, and "False" otherwise.""" \\

\textbf{2. Reformat complex reasoning:} \\

"""<Thought Process> \\
\{\} \\
</Thought Process> \\

<Question> \\
\{\} \\
</Question> \\

The <Thought Process> above reflects the model's reasoning based on the <Question>. Your task is to rewrite the <Thought Process> to resemble a more human-like, intuitive natural thinking process. The new version should: \\

1. Be presented as step-by-step reasoning, with each thought on a new line separated by a line break. \\
2. Avoid structured titles or formatting, or phrasing it as something an agent said, focusing on natural transitions. Use casual and natural language for transitions or validations, such as "hmm," "oh," "also," or "wait." \\
3. Expand the content, making the reasoning richer, more detailed, and logically clear while still being conversational and intuitive. \\

Return directly the revised natural thinking in JSON format as follows: \\
```json \\
\{\{ \\
  "NaturalReasoning": "..." \\
\}\} \\
```""" \\

\textbf{3. Prompt for obtaining the final answer:} \\

"""<Internal Thinking> \\
\{\} \\
</Internal Thinking> \\

<Question> \\
\{\} \\
</Question> \\

The <Internal Thinking> represents your internal thoughts about the <Question>. Based on this, generate a rich and high-quality final response to the user. If there is a clear answer, provide it first. Ensure your final response closely follows the <Question>. The response style should resemble GPT-4's style as much as possible. Output only your final response, without any additional content.""" \\

\end{minipage}
}
\end{figure*}

\end{document}